\documentclass[journal]{IEEEtran}
\usepackage{amssymb}
\usepackage[algoruled,vlined,linesnumbered]{algorithm2e}
\usepackage{comment}
\usepackage{amsthm}

\theoremstyle{definition}

\theoremstyle{remark}

\usepackage{bm}
\usepackage{cite} 
\usepackage{amsmath}
\usepackage{float}
\usepackage{graphicx}
\usepackage{subfigure}
\usepackage{xcolor}
\usepackage{array}
\usepackage{mathtools}
\usepackage{tabularx}

\newcolumntype{C}[1]{>{\centering\let\newline\\\arraybackslash\hspace{0pt}}m{#1}}

\usepackage[colorinlistoftodos]{todonotes}

\begin{document}
\title{Learning Deep Analysis Dictionaries--Part II: Convolutional Dictionaries}

\author{Jun-Jie~Huang,~\IEEEmembership{Student Member,~IEEE,} and~Pier
Luigi~Dragotti,~\IEEEmembership{Fellow,~IEEE}}

\maketitle

\begin{abstract}

In this paper, we introduce a Deep Convolutional Analysis Dictionary Model (DeepCAM) by learning convolutional dictionaries instead of unstructured dictionaries as in the case of deep analysis dictionary model introduced in the companion paper. 
Convolutional dictionaries are more suitable for processing high-dimensional signals like for example images and have only a small number of free parameters. By exploiting the properties of a convolutional dictionary, we present an efficient convolutional analysis dictionary learning approach. 
A $L$-layer DeepCAM consists of $L$ layers of convolutional analysis dictionary and element-wise soft-thresholding pairs and a single layer of convolutional synthesis dictionary.
Similar to DeepAM, each convolutional analysis dictionary is composed of a convolutional Information Preserving Analysis Dictionary (IPAD) and a convolutional Clustering Analysis Dictionary (CAD). 
The IPAD and the CAD are learned using variations of the proposed learning algorithm. 
We demonstrate that DeepCAM is an effective multi-layer convolutional model and, on single image super-resolution, achieves performance comparable with other methods while also showing good generalization capabilities.

\end{abstract}


\IEEEpeerreviewmaketitle

\section{Introduction}

\IEEEPARstart{C}{onvolutional} dictionary learning has attracted increasing interests in signal and image processing communities
as it leads to a more elegant framework for high-dimensional signal analysis.
An advantage of convolutional dictionaries \cite{bristow2013fast, wohlberg2014efficient, heide2015fast, pfister2018learning, chun2017convolutional,chun2019convolutional, cai2014data, papyan2017working, wohlberg2015efficient, papyan2017convolutional, papyan2016multi,sulam2017multi,aberdam2019multi} is that they can take the high-dimensional signal as input for sparse representation and processing, whereas traditional approaches \cite{ksvd2006,elad2006image,yang2010image,zeyde2010single,ravishankar2012learning,rubinstein2013analysis,hawe2013analysis} have to divide the high-dimensional signal into overlapping low-dimensional patches and perform sparse representation on each patch independently.

A convolutional dictionary models the convolution between a set of filters and a signal. It is a structured dictionary and can be represented as a concatenation of Toeplitz matrices where each Toeplitz matrix is constructed using the taps of a filter and the usual assumption is that the filters are with compact support. So a convolutional dictionary is effective for processing high-dimensional signals while also restraining the number of free parameters.

To achieve efficient convolutional dictionary learning, the convolutional dictionary is usually modelled as a concatenation of circulant matrices \cite{bristow2013fast, wohlberg2014efficient, heide2015fast,pfister2018learning,chun2017convolutional,chun2019convolutional} by assuming a periodic boundary condition on the signals. As all circulant matrices share the same set of eigenvectors which is the Discrete Fourier Transform (DFT) matrix, a circular convolution can be therefore represented as a multiplication in Fourier domain and can be efficiently implemented using Fast Fourier Transform (FFT). 
However, using a circulant matrix to approximate a general Toeplitz matrix may lead to boundary artifacts \cite{heide2015fast,kavukcuoglu2010learning,chun2018convolutional} especially when the boundary region is large.


A multi-layer convolutional dictionary model is able to represent multiple levels of abstraction of the input signal.
Due to the associativity property of convolution, multiplying two convolutional dictionaries results in a convolutional dictionary whose corresponding filters are the convolution of two set of filters and have support size that is larger than the original filters. 
In the Multi-Layer Convolutional Sparse Coding (ML-CSC) model in \cite{papyan2016multi,sulam2017multi,papyan2018theoretical}, there are multiple layers of convolutional synthesis dictionaries. 
By increasing the number of layers, the global dictionary which is the multiplication of the convolutional dictionaries is able to represent more complex structures. 
The ML-CSC model \cite{papyan2016multi} provides theoretical insights on the conditions for the success of the layered sparse pursuit and uses it as a way to interpret the forward pass of Deep Neural Networks (DNNs) as a layered sparse pursuit.

Convolutional Neural Networks (CNNs) \cite{lecun1990handwritten, lecun1998gradient, simonyan2014very} have been widely used for processing images and achieve state-of-the-art performances in many applications. A CNN consists of a cascade of convolution operations and element-wise non-linear operations. The Rectified Linear Unit (ReLU) activation function is one of the most popular non-linear operators used in DNNs. With the multi-layer convolution structure, a deeper layer in a CNN receives information from a corresponding wider region of the input signal. The ReLU operator provides a non-linear transformation and leads to a sparse feature representation. The parameters of a CNN are usually optimized using the backpropagation algorithm \cite{rumelhart1985learning} with stochastic gradient descent. 

In this paper, we propose a Deep Convolutional Analysis Dictionary Model (DeepCAM) which consists of multiple layers of convolutional analysis dictionaries and soft-thresholding operations and a layer of convolutional synthesis dictionary. 
The motivation is to use DeepCAM as a tool to interpret the workings of CNNs from the sparse representation perspective.
Single image super-resolution \cite{yang2010image, zeyde2010single, timofte2013anchored, timofte2014a+, huang2015practical, huang1215_learning, huang2015fast, huang0717_SRHRF+, dong2015image, dong2016image, dong2016accelerating, shi2016real} is used as a sample application to validate the proposed model design.
The input low-resolution image is used as the input to DeepCAM and is not partitioned into patches.
At each layer of DeepCAM, the input signal is multiplied with a convolutional analysis dictionary and then passed through soft-thresholding operators. At the last layer, the convolutional synthesis dictionary is used to predict the high-resolution image.

The contribution of this paper is two-fold:
\begin{itemize}

  \item We propose a Deep Convolutional Analysis Dictionary Model (DeepCAM) for single image super-resolution, where 
  at each layer, the convolutional analysis dictionary and the soft-thresholding operation are designed to achieve simultaneously information preservation and discriminative representation. 
  
  \item We propose a convolutional analysis dictionary learning method by explicitly modelling the convolutional dictionary with a Toeplitz structure. By exploiting the properties of Toeplitz matrices, the convolutional analysis dictionary can be efficiently learned from a set of training samples. Simulation results on single image super-resolution are used to validate our proposed DeepCAM and convolutional dictionary learning method.

\end{itemize}

The rest of the paper is organized as follows. 
In Section II we show how we build a convolutional analysis dictionary from an unstructured dictionary. 
In Section III, we propose an efficient convolutional analysis dictionary learning algorithm by exploiting the properties of Toeplitz matrices. 
Section IV presents the proposed Deep Convolutional Analysis Dictionary Model (DeepCAM) and the complete learning algorithm.
Section V presents simulation results on single image super-resolution task and Section VI draws conclusions.

\section{Convolutional Analysis Dictionary}
\label{ConvAD}

An analysis dictionary $\bm{\Omega} \in \mathbb{R}^{m \times n}$ contains $m$ row atoms $\{ \bm{\omega}_i^T \in \mathbb{R}^{n} \}_{i=1}^m$ and is usually assumed to be over-complete with $m \geq n$. Given a signal of interests $\bm{\alpha} \in \mathbb{R}^{n}$, the analysis dictionary $\bm{\Omega}$ should be able to sparsify $\bm{\alpha}$ while preserving its essential information. That is, the analysis coefficients $\bm{\Omega} \bm{\alpha}$ are sparse but still contain sufficient information for further processing. 

The focus of this paper is on learning convolutional analysis dictionaries which model the convolution between a signal and a set of filters. The filters' taps depend on the rows of the analysis dictionary. In what follows, we first show how we build our convolutional analysis dictionary from a unstructured analysis dictionary. We then study strategies to learn a convolutional analysis dictionary from a set of training samples.

The convolution can be represented as a multiplication with a convolutional analysis dictionary. 
Let us assume that the input signal is 1-dimensional for simplicity. The convolution between an atom $\bm{\omega}_i \in \mathbb{R}^{n}$ and an input signal $\bm{x}\in \mathbb{R}^l$ with $l>n$ can be expressed as:
\begin{equation}
    \bm{\omega}_i * \bm{x} = \mathbf{T}(\bm{\omega}_i^T,l)\bm{x},
\end{equation}
where $*$ denotes the convolution operator, and $\mathbf{T}(\bm{\omega}_i^T,l)$ is a Toeplitz matrix with $l$ columns which is constructed using $\bm{\omega}_i$ as follows:
\begin{equation}
    \mathbf{T}(\bm{\omega}_i^T,l) = \sum_{j=1}^n \bm{\omega}_i (j) \mathbf{T}_j,
    \label{eq:Toep1D}
\end{equation}
where $\bm{\omega}_i (j)$ is the $j$-th coefficient of $\bm{\omega}_i$, and $\mathbf{T}_j \in \mathbb{R}^{p \times l}$ with $p=l-n+1$ is an indicator matrix with 1s on the $j$-th upper diagonal and 0s on other locations.

Given an unstructured analysis dictionary $\bm{\Omega} \in \mathbb{R}^{m \times n}$ and an input signal $\bm{x}\in \mathbb{R}^l$, the convolution between $\bm{x}$ and each row of $\bm{\Omega}$ can be expressed as a matrix multiplication where $\bm{\Omega}$ is converted to a convolutional analysis dictionary $\bm{\mathcal{H}}(\bm{\Omega}, l) \in \mathbb{R}^{pm \times l}$ which can be represented as a concatenation of $m$ Toeplitz matrices along the column direction:
\begin{equation}
    \bm{\mathcal{H}}(\bm{\Omega}, l) = [\mathbf{T}(\bm{\omega}_1^T,l);\cdots;\mathbf{T}(\bm{\omega}_m^T,l)].
\end{equation}

\begin{figure}[t]
    \centering
	\hspace*{\fill}
	\subfigure[An analysis dictionary $\bm{\Omega}$.]{
		\label{fig:conv1} 
		\includegraphics[width=0.18\textwidth]{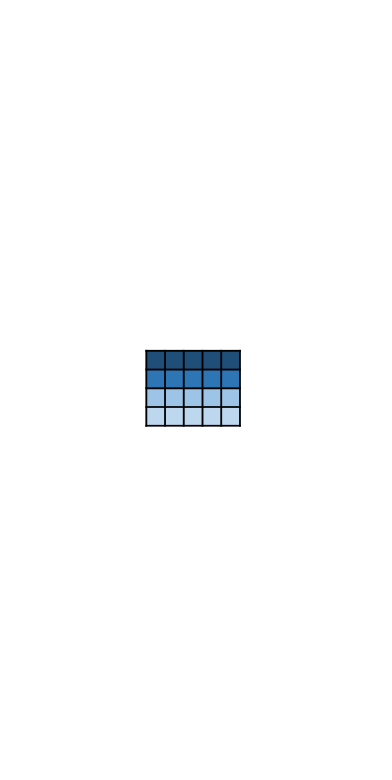}}
		\hfill
	\subfigure[The convolutional analysis dictionary $\bm{\mathcal{H}}(\bm{\Omega},l)$.]{
		\label{fig:conv2} 
		\includegraphics[width=0.18\textwidth]{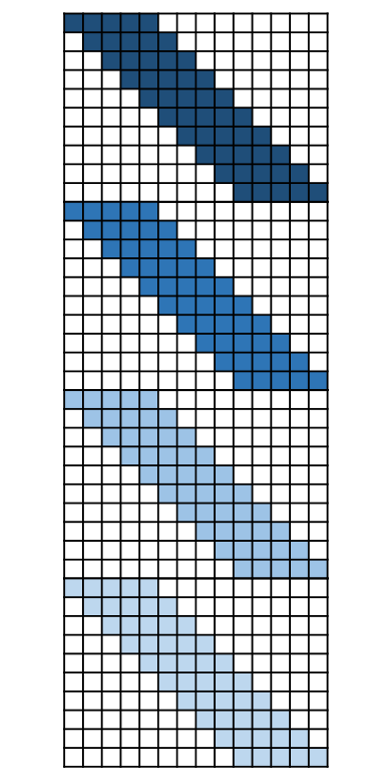}}
	\hspace*{\fill}
	\caption{A convolutional analysis dictionary $\bm{\mathcal{H}}(\bm{\Omega},l)$  with $l=12$ is a concatenation of $m=4$ Toeplitz matrices $\{ \mathbf{T}(\bm{\omega}_i^T,l) \}_{i=1}^{m}$.}
    \label{fig:ConvDict}
\end{figure}

Instead of assuming a circulant structure, we model the convolutional operation with a Toeplitz matrix. The represented convolution operation will be performed only within the input signal. That is, convolutional operation is performed without padding at the boundaries. 

Fig. \ref{fig:ConvDict} shows an example of how we build a convolutional analysis dictionary $\bm{\mathcal{H}}(\bm{\Omega},l)$ 
from the unstructured analysis dictionary $\bm{\Omega} \in \mathbb{R}^{m \times n}$ with $m=4, n=5$ and $l=12$. Note that the analysis dictionary $\bm{\Omega}$ in Fig. \ref{fig:conv1} is not over-complete, while the convolutional analysis dictionary $\bm{\mathcal{H}}(\bm{\Omega},l)$ has more rows than columns.

\begin{figure}[t]
    \centering
	\hspace*{\fill}
	\subfigure[The convolution operation between a $6 \times 6$ image region and a $3 \times 3$ filter.]{
		\label{fig:conv2D_1} 
		\includegraphics[width=0.25\textwidth]{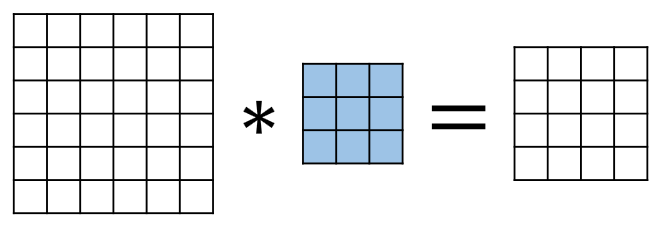}}
	\hspace*{\fill}

	\hspace*{\fill}
	\subfigure[The corresponding Toeplitz matrix.]{
		\label{fig:conv2D_2} 
		\includegraphics[width=0.3\textwidth]{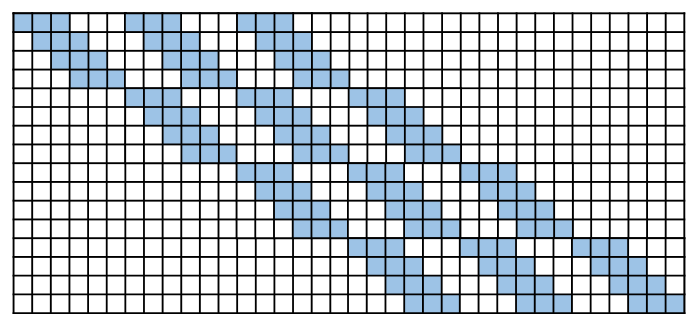}}
	\hspace*{\fill}
    \caption{An example of convolution between a 2-dimensional convolutional filter with 2-dimensional data and the convolution can be represented by a doubly block Toeplitz matrix.}
    \label{fig:2DConvDict}
\end{figure}

The above description applies to 1-dimensional input signals, but it can be extended to multi-dimensional signals, like for example images. 
A convolutional analysis dictionary will then be in the form of a concatenation of doubly block Toeplitz matrices (i.e. a matrix with block Toeplitz structure where each block is a Toeplitz matrix). Similar to Eqn. (\ref{eq:Toep1D}), the doubly block Toeplitz structure can be represented with corresponding indicator matrices.
Fig. \ref{fig:conv2D_1} shows an example of 2-dimensional convolution between a $6 \times 6$ image region and a $3 \times 3$ filter and Fig. \ref{fig:conv2D_2} shows the corresponding convolutional analysis dictionary.

\section{Learning Convolutional Analysis Dictionaries}
\label{learnConvDict}

In this section, we propose an efficient convolutional analysis dictionary learning algorithm by exploiting the properties of the Toeplitz structure within the dictionary.

For simplicity, let us assume that the input data is made of 1-dimensional vectors. Therefore, the convolutional analysis dictionary is a concatenation of Toeplitz matrices as we discussed in Section \ref{ConvAD}. The proposed learning algorithm can be easily extended to the multi-dimensional case where the convolutional dictionary is a concatenation of doubly block Toeplitz matrices.

Let us assume that convolution is performed between an analysis dictionary $\bm{\Omega} \in \mathbb{R}^{m \times n}$ and an input signal $\bm{x} \in \mathbb{R}^{l}$ with $l>n$.
Therefore, the convolutional analysis dictionary $\bm{\mathcal{H}}(\bm{\Omega},l)$ will be of size $q \times l$ with $q = pm$.
Given that $\bm{\mathcal{H}}(\bm{\Omega},l)$ is built using $\bm{\Omega}$, it has the same number of free parameters as $\bm{\Omega}$ despite being a much bigger matrix. This means that if we were to optimize $\bm{\mathcal{H}}(\bm{\Omega},l)$ directly we would end up with a computationally inefficient approach.

\begin{figure}
    \centering
    \includegraphics[width=0.25\textwidth]{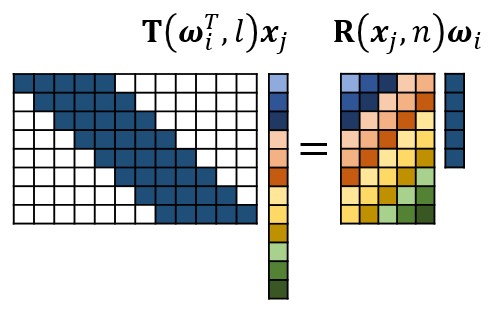}
    \caption{Convolution can be represented as multiplying with a Toeplitz matrix or multiplying with a right dual matrix. The right dual matrix is not sparse.}
    \label{fig:Rightdual}
\end{figure}

We mitigate this issue by first observing that when we assume that the convolutional filters are with compact support $n \ll l$, the convolutional analysis dictionary $\bm{\mathcal{H}}(\bm{\Omega},l)$ has many zero entries. It is therefore inefficient to evaluate $\bm{\mathcal{H}}(\bm{\Omega},l)\bm{x}$ by directly multiplying $\bm{\mathcal{H}}(\bm{\Omega},l)$ with $\bm{x}$.
As illustrated in Fig. \ref{fig:Rightdual}, with the commutativity property of convolution (i.e. $\bm{a} * \bm{b} = \bm{b} * \bm{a}$), the matrix multiplication between a Toeplitz matrix $ \mathbf{T}(\bm{\omega}_i^T,l) \in \mathbb{R}^{p \times l}$ and an input vector $\bm{x} \in \mathbb{R}^l$ can be efficiently implemented as:
\begin{equation}
    \mathbf{T}(\bm{\omega}_i^T,l)\bm{x} = \mathbf{R}(\bm{x}_j,n)\bm{\omega}_{i},
    \label{eq:convEquiv}
\end{equation}
where $\mathbf{R}(\bm{x},n) \in \mathbb{R}^{p \times n}$ is the right dual matrix of $\mathbf{T}(\bm{\omega}_i^T,l)$ and is defined as: 
\begin{equation}
    \mathbf{R}(\bm{x},n) = \sum_{j=1}^l \bm{x}(j) \mathbf{R}_j,
\end{equation}
where $\bm{x}(j)$ is the $j$-th coefficient of $\bm{x}$, and $ \mathbf{R}_j \in \mathbb{R}^{p \times n}$ is an indicator matrix with 1s on the $j$-th skew-diagonal and 0s on other locations.
Note that the right dual matrix $\mathbf{R}(\bm{x},n)$ is a matrix without zero entries and has $n \ll l$ columns.

\begin{table*}[]
    \centering
    \begin{tabular}{|c||C{1.3cm}|C{1.3cm}|C{1.3cm}|C{1.3cm}|C{1.3cm}|C{1.3cm}|C{1.3cm}|C{1.3cm}|C{1.3cm}|}
        \hline 
        Symbol     & $\bm{\Omega}$ & $\bm{\mathcal{H}}(\bm{\Omega},l)$  & $\mathbf{T}(\bm{\omega}_i^T,l)$  & $\mathbf{R}(\bm{w}_{i},n)$ & $\bm{W}$ & $\bm{U}$ & $\bm{Q}$ & $\bm{P}_S$ & $\bm{P}_{\bm{\omega}}$ \\ 
        \hline 
        Size & $m \times n$ & $pm \times l$ & $ p \times l$ & $p \times n$ & $l \times K$ & $l \times (l-K)$ & $pm \times K$ & $l \times l$ & $n \times n$ \\
        \hline 
    \end{tabular}
    \caption{A list of symbols and their dimensions. For simplicity, in the table we denote $p = l - n + 1$ with $l \gg n$.}
    \label{tab:symbols}
\end{table*}

Secondly, whenever possible, we will pose the optimization problem using $\bm{\Omega}$ whilst imposing the constraints associated with the structured matrix $\bm{\mathcal{H}}(\bm{\Omega},l)$ as the actual analysis dictionary learning problem.

We want the convolutional analysis dictionary $\bm{\mathcal{H}}(\bm{\Omega},l)$ to satisfy four properties: (i) its row atoms span the input data space; (ii) it is able to sparsify the input data; (iii) the row atoms are of unit norm; (iv) there are no pairs of row atoms in $\bm{\Omega}$ that are linearly dependent.

Different from the unstructured analysis dictionary learning case, we propose to use two sets of input training data with different sizes.
Let us denote the super-patch training data as $\bm{\mathcal{X}}_S=[\bm{x}_{S1}, \bm{x}_{S2}, \cdots, \bm{x}_{SN_1}] \in \mathbb{R}^{l \times N_1}$ and denote the small patch training data as $\bm{\mathcal{X}}=[\bm{x}_{1}, \bm{x}_{2}, \cdots, \bm{x}_{N_2}] \in \mathbb{R}^{n \times N_2}$. Please remember that $n$ is the filters' size and $l$ is the dimension of input signals and we assume $n < l$. Both the super-patch and small patch training datasets are extracted from an external training dataset. The super-patch training data will be used to impose property (i) which is a global property of the convolutional dictionary. The small patch training data will be used to impose property (ii).

The first learning objective is that the convolutional analysis dictionary $\bm{\mathcal{H}}(\bm{\Omega},l)$ should be able to span the input data space in order to preserve the information within the input super-patch training data $\bm{\mathcal{X}}_S$. 
The super-patch training dataset defines the subspace covered by the input data. Let us denote with $\bm{W} \in \mathbb{R}^{l \times K}$ the orthogonal basis covering the signal subspace of the input super-patch data $\bm{\mathcal{X}}_S$ where we assume that this subspace has dimension $K$, we also denote with $\bm{U}\in\mathbb{R}^{l \times (l - K)}$ the orthogonal basis of the orthogonal complement to the signal subspace of $\bm{\mathcal{X}}_S$. These two bases will be used to impose that the row space of the learned convolutional analysis dictionary spans the input data space while being orthogonal to the null-space of $\bm{\mathcal{X}}_S$. 
The information preservation constraint can be interpreted as a rank constraint on the convolutional analysis dictionary which is usually achieved by imposing a logarithm determinant constraint:
\begin{equation}
    \begin{split}
        &h(\bm{\mathcal{H}}(\bm{\Omega},l)) = \\
        &- \frac{1}{K \log K} \log \det \left(\frac{1}{q} \bm{W}^T \bm{\mathcal{H}}(\bm{\Omega},l)^T \bm{\mathcal{H}}(\bm{\Omega},l) \bm{W} \right).
    \end{split}
    \label{eq:log-det}
\end{equation}

The size of the convolutional analysis dictionary can be huge, especially when the input data is multi-dimensional. Therefore it would be computationally inefficient to evaluate Eqn. (\ref{eq:log-det}) and its first order derivative directly.
By exploiting the properties of the convolutional analysis dictionary, we propose an efficient reformulation of Eqn. (\ref{eq:log-det}) which is based on the analysis dictionary $\bm{\Omega}$. 


With the definition of the right dual matrix, the multiplication between $\bm{\mathcal{H}}(\bm{\Omega},l)$ and the $j$-th orthogonal basis element $\bm{w}_{j}$ of $\bm{W}$ can be expressed as:
\begin{equation}
    \bm{\mathcal{H}}(\bm{\Omega},l)\bm{w}_{j}= \text{vec}\left( \mathbf{R}(\bm{w}_{j},n) \bm{\Omega}^T \right),
\end{equation}
where $\text{vec}(\cdot)$ denotes the vectorization operation $\text{vec}(\bm{A}): \mathbb{R}^{m_1 \times \cdots \times m_{D_a}} \rightarrow \mathbb{R}^{\prod_{k=1}^{D_a} m_k}$, the vectorization operation for 2-dimensional signal can be expressed as:
\begin{equation}
    \text{vec}(\bm{A}) = \sum_{i=1}^n \bm{e}_i \otimes \bm{A} \bm{e}_i,
\end{equation}
where $\bm{e}_i$ is the $i$-th canonical basis vector of $\mathbb{R}^n$, that is, $\bm{e}_i = [0, \cdots, 0, 1, 0, \cdots, 0]^T \in \mathbb{R}^n$ (with 1 on the $i$-th location), and $\otimes$ denotes the Kroneckers product.

The information preservation constraint in Eqn. (\ref{eq:log-det}) can therefore be reformulated and expressed in terms of the analysis dictionary $\bm{\Omega}$ as:
\begin{equation}
     h(\bm{\Omega}) = - \frac{1}{K \log K} \log \det \left(\frac{1}{q} \bm{Q}^T \bm{Q}  \right),
    \label{eq:log-det2}
\end{equation}
where $\bm{Q} = \left[ \bm{q}_i,\cdots,\bm{q}_K \right]$ with $\bm{q}_{i} = \text{vec}(\mathbf{R}(\bm{w}_{i},n)\bm{\Omega}^T)$.

The gradient of $h(\bm{\Omega})$ can be expressed as:
\begin{equation}
    \frac{\partial}{\partial \bm{\Omega}} h(\bm{\Omega}) = - \frac{2}{K \log K} \left(\sum_{i=1}^K \bm{\Omega} \bm{\Sigma}_{i}  \right) \bm{\Sigma}_{Q}^{-1},
\end{equation}
where $\bm{\Sigma}_{i}= \mathbf{R}(\bm{w}_{i},n)^T \mathbf{R}(\bm{w}_{i},n)$ and $\bm{\Sigma}_{Q} =  \bm{Q}^T \bm{Q}$.

With the information preservation constraint in Eqn. (\ref{eq:log-det2}), the learned $\bm{\mathcal{H}}(\bm{\Omega},l)$ is constrained to span the signal subspace defined by $\bm{W}$. However, we still need to exclude the null-space components of the training data from $\bm{\mathcal{H}}(\bm{\Omega},l)$.
Specifically, the Toeplitz matrix $\mathbf{T}(\bm{\omega}_i^T,l)$ should not be within the subspace spanned by $\bm{U}$ to avoid a zero response when multiplying with $\bm{\mathcal{X}}_S$. 

Therefore we define the feasible set of the convolutional analysis dictionary $\bm{\mathcal{H}}(\bm{\Omega},l)$ as $\Theta_{{\mathcal{H}}}=\mathbb{S}_{l-1}^{q}\cap\bm{U}^{\perp}$ with $\mathbb{S}_{l-1}$ being the unit sphere in $\mathbb{R}^{l}$, $\mathbb{S}_{l-1}^{q}$ being the product of $q$ unit spheres $\mathbb{S}_{l-1}$ and $\bm{U}^{\perp}$ being the orthogonal complement of $\bm{U}$. The unit sphere constraint ensures that the unit norm condition is satisfied. The feasible set $\Theta_{{\mathcal{H}}}$ is defined in $\mathbb{R}^l$, while we wish to have a feasible set for $\bm{\Omega}$ which is defined in $\mathbb{R}^n$ with $n \ll l$ and can be more efficiently implemented.

The operation of orthogonal projection onto the complementary subspace of $\bm{U}$ can be represented by the projection matrix $\bm{P}_S \in \mathbb{R}^{l \times l}$ given by:
\begin{equation}
    \bm{P}_S=\mathbf{I}_l-{\bm{U}^T}^{\dagger} \bm{U}^{T},
\end{equation}
where $\mathbf{I}_l\in\mathbb{R}^{l \times l}$ is the identity matrix and ${\bm{U}^T}^{\dagger}$ is the pseudo-inverse of $\bm{U}^{T}$.

The orthogonal projection operation is achieved by multiplying the convolutional analysis dictionary with the projection matrix. The projection is applied on the rows of $ \bm{\mathcal{H}}(\bm{\Omega},l)$.
With the definition of the right dual matrix, the orthogonal projection operation can be expressed in terms of the analysis dictionary atom $\bm{\omega}_i^T$ as:
\begin{equation}
   \bm{P}_S \mathbf{T}(\bm{\omega}_i^T,l)^T = \left[\mathbf{R}(\bm{p}_{S1},n)\bm{\omega}_{i}, \cdots, \mathbf{R}(\bm{p}_{Sl},n)\bm{\omega}_{i}\right]^T,
\end{equation}
where $\bm{p}_{Sj}^T$ denotes the $j$-th row of $\bm{P}_S$.

We note that, after the projection, the Toeplitz structure within $\mathbf{T}(\bm{\omega}_i^T,l)\bm{P}_S^T$ may not be preserved and needs to be imposed again. The Toeplitz matrix closest to $\mathbf{T}(\bm{\omega}_i^T,l)\bm{P}_S^T$ is obtained by averaging over the diagonal elements \cite{cadzow1988signal}. 
The orthogonal projection operation and the averaging operation can be jointly represented and applied to the atoms of the analysis dictionary.
Let us define a vector $\bm{p}_j$ whose inner product with $\bm{\omega}_i$ equals the average value of the $j$-th diagonal elements of $\mathbf{T}(\bm{\omega}_i^T,l)\bm{P}_S^T$, it can be expressed as:
\begin{equation}
    \bm{p}_j^T = \frac{1}{n} \sum_{k=j}^{j+n-1} \bm{r}_{Sk,k-j+1}^T,
\end{equation}
where $\bm{r}_{Sk,i}^T$ denotes the $i$-th row of $\mathbf{R}(\bm{p}_{Sk},n)$.

The matrix $\bm{P}= \left[\bm{p}_1, \cdots, \bm{p}_n \right]^T \in \mathbb{R}^{n \times n}$ therefore represents simultaneously the orthogonal projection operation and the averaging operation. 
Let us denote the feasible set of the analysis dictionary $\bm{\Omega}$ as $\Theta_{{\Omega}}=\mathbb{S}_{n-1}^{m}\cap\bm{V}^{\perp}$ where $\mathbb{S}_{n-1}^{m}$ represents the product of $m$ unit spheres $\mathbb{S}_{n-1}$ and $\bm{V}^{\perp}$ represents the orthogonal complementary subspace of the null-space of $\bm{\mathcal{X}}$. 
The operation of the orthogonal projection onto the tangent space $\mathcal{T}_{\bm{\bm{\omega}}}(\Theta_{{\Omega}})$ can then be represented by the projection matrix $\bm{P_{\omega}}$:
\begin{equation}
    \bm{P_{\omega}}=\bm{P} \left(\mathbf{I}_n-\bm{Q}_{\bm{\omega}}^{\dagger}\bm{Q}_{\bm{\omega}} \right),
    \label{eq:projMat}
\end{equation}
where $\mathbf{I}_n\in\mathbb{R}^{n\times n}$ is the identity
matrix, and $\bm{Q}_{\bm{\omega}}=[2\bm{\omega},\bm{V}]^{T}\in\mathbb{R}^{(n-k+1)\times n}$.

The sparsifying property of the convolutional analysis dictionary $\bm{\mathcal{H}}(\bm{\Omega},l)$ over the super-patch training data can be achieved by imposing the sparsifying property of the analysis dictionary $\bm{\Omega}$ over the small patch training data. The rationale is that the row atoms of the convolutional analysis dictionary only operate on local regions of the input signal as illustrated in Eqn. (\ref{eq:convEquiv}) and Fig. \ref{fig:Rightdual}.
Similar to \cite{hawe2013analysis,kiechle2015bimodal,huang2019DeepAM}, the sparsifying constraint is imposed by using a log-square function which promotes analysis dictionary atoms that sparsify the small patch training data:
\begin{equation}\label{eq:sparsify}
    g(\bm{\Omega}) = \frac{1}{N_2 m \log(1+\nu)}\sum_{i=1}^{N_2} \sum_{j=1}^{m}\log \left(1 + \nu (\bm{\omega}_{j}^T \bm{x}_{i})^2 \right),
\end{equation}
where $\nu$ is a tunable parameter which controls the sparsifying ability of the learned dictionary.

The linearly dependent penalty and the unit norm constraint can also be imposed directly on the analysis dictionary $\bm{\Omega}$.
Linearly dependent row atoms (e.g. $\bm{\omega}_{i}^{T}=\pm\bm{\omega}_{j}^{T}$) are penalized by using a logarithm barrier term $l\left(\bm{\Omega}\right)$:
\begin{equation}
    l\left(\bm{\Omega}\right)=-\frac{1}{m\left(m-1\right)}\underset{1\leq i<j\leq m}{\sum}\log\left(1-\left(\bm{\omega}_{i}^{T}\bm{\bm{\omega}}_{j}\right)^{2}\right).\label{eq:log-barrier}
\end{equation}

We observe that, by exploiting the Toeplitz structure, we have been able to impose the desired proprieties of a convolutional analysis dictionary by imposing constraints on the lower-dimensional analysis dictionary $\bm{\Omega}$. This will reduce computational costs and memory requirements.

Combining the information preservation constraint in Eqn. (\ref{eq:log-det2}), feasible set constraint in Eqn. (\ref{eq:projMat}), sparsifying constraint in Eqn. (\ref{eq:sparsify}), and linearly dependent penalty term in Eqn. (\ref{eq:log-barrier}), the objective function for the convolutional analysis dictionary problem can be expressed as:
\begin{equation}
    \bm{\Omega}=\arg\underset{\bm{\Omega}^{T}\in\Theta_{\bm{\Omega}}}{\min}f(\bm{\Omega}),
    \label{eq:GOAL+}
\end{equation}
where $f(\bm{\Omega})=g(\bm{\Omega})+\kappa h(\bm{\Omega})+\upsilon l(\bm{\Omega})$ with $\kappa$ and $\upsilon$ being the regularization parameters.

The proposed convolutional analysis dictionary learning algorithm ConvGOAL+ is summarized in \textbf{Algorithm 1}. 
The objective function in Eqn. (\ref{eq:GOAL+}) is optimized using a geometric conjugate gradient descent method (see also \cite{absil2009optimization,hawe2013analysis,huang2019DeepAM}).

\begin{algorithm}[t]
    \SetAlgoLined
    \textbf{Input:} number of filters $m$, support size $n$, training data $\bm{\mathcal{X}}_S \in \mathbb{R}^{l \times N_1}$ and $\bm{\mathcal{X}} \in \mathbb{R}^{n \times N_2}$;
    
    \textbf{Initialize:} Initialized $\bm{\Omega}^{(0)} \in \mathbb{R}^{m \times n}$, $t=0$\;
    
    Find the orthogonal basis $\bm{W}$ and $\bm{U}$ for the input data subspace of  $\bm{\mathcal{X}}_S$ and its orthogonal complementary subspace, respectively;
    
    Find the orthogonal basis $\bm{V}$ for the orthogonal complementary subspace of  $\bm{\mathcal{X}}$;
    
    \While{halting criterion false}{
        $t \leftarrow t + 1$ \;
         
        Compute gradient of the objective function $\nabla f(\bm{\Omega}^{(t)})$;

        Orthogonal project $\nabla f(\bm{\Omega}^{(t)})$ onto the tangent space of manifold $\Theta_{{\Omega}}$ at $\bm{\Omega}^{(t)}$;

        Update $\bm{\Omega}^{(t+1)}$ along the search direction using backtracking line search.
    }
    \textbf{Output:} Learned analysis dictionary $\bm{\Omega}$.
 \caption{ConvGOAL+ Algorithm}
\end{algorithm}

\section{Deep Convolutional Analysis Dictionary Model}

\begin{figure*}[ht]
    \centering
    \includegraphics[width=0.95\linewidth]{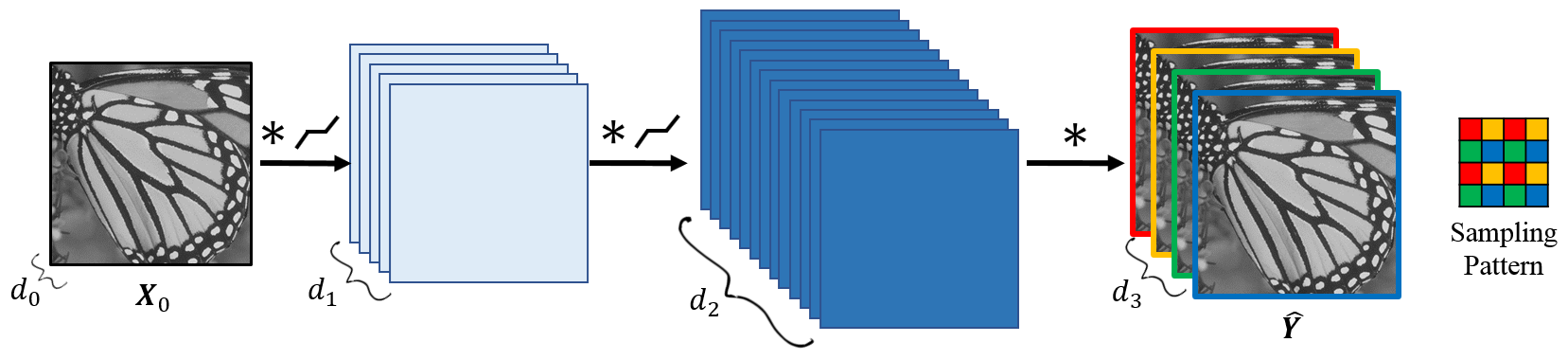}
    \caption{A 2-layer Deep Convolutional Analysis dictionary Model for $2 \times$ single image super-resolution.
    There are 2 layers of analysis dictionaries $\left\{ \bm{\Omega}_{i}\right\} _{i=1}^{2}$ with element-wise soft-thresholding
    operators $\left\{ \mathcal{S}_{\bm{\lambda}_{i}}\left(\cdot\right)\right\} _{i=1}^{2}$ and a layer of synthesis dictionary $\bm{D}$. The input image is a gray image. The estimated $d_3 = 4$ HR images $\widehat{\bm{Y}}$ is obtained through a cascade of convolution and soft-thresholding operations with input LR image $\bm{X}_{0}$. The final predicted HR image is obtained by reshaping $\widehat{\bm{Y}}$ according to the sampling pattern.}
    \label{fig:overall}
\end{figure*}

In this section, we introduce our Deep Convolutional Analysis Dictionary Model (DeepCAM).
DeepCAM is a convolutional extension of the Deep Analysis dictionary Model (DeepAM) \cite{huang2019DeepAM}. 
Different from DeepAM which is patch-based,
DeepCAM performs convolution operation and element-wise soft-thresholding at image level on all layers without dividing the input image into patches.

When it comes to Single Image Super-Resolution (SISR), convolutional neural networks are designed using two main strategies: the early-upsampling approaches \cite{dong2015image, dong2016image} and the late-upsampling approaches \cite{dong2016accelerating, shi2016real}. The early-upsampling approaches \cite{dong2015image, dong2016image} first upsample the low-resolution (LR) image to the same resolution of the desired high-resolution (HR) one through bicubic interpolation and then perform convolution on the upsampled image. The drawback is that this leads to a large number of model parameters and a high computational complexity during testing as the feature maps are of the same size as the HR image. The late-upsampling approaches \cite{dong2016accelerating, shi2016real} perform convolution on the input LR image and applies a deconvolution layer \cite{dong2016accelerating} or a sub-pixel convolution layer \cite{shi2016real} at the last layer to predict the HR image. The late-upsampling approaches have smaller number of parameters and lower computational cost than the early-upsampling one.

SISR is used as a sample application to validate our proposed design.
We utilize a similar strategy as the late-upsampling approach. The LR image is used as input to DeepCAM without bicubic interpolation. At each layer, the convolution and soft-thresholding operations are applied to the corresponding input signal. For SISR with up-sampling factor $s$, the synthesis dictionary consists of $s^2$ atoms. The convolution between the synthesis dictionary and its input signal yields $s^2$ output channels which correspond to $s^2$ sub-sampled version of the HR image. The final predicted HR image can then be obtained by reshaping and combing the $s^2$ output channels.

The parameters of a $L$-layer DeepCAM include $L$ layers of analysis dictionary and soft-thresholds pair $\{ (\bm{\Omega}_i, \bm{\lambda}_i) \}_{i=1}^L$ and a single synthesis layer modelled with dictionary $\bm{D}$. The atoms of the dictionaries represent filters. Let us denote with $d_i$ the number of filters at the $i$-th layer and with $p_i \times p_i$ the spatial support size of the convolutional filters, since there were $d_{i-1}$ filters at the previous layer, there are $d_i \times n_i$ free parameters at layer $i$ with $n_i = p_i^2 d_{i-1}$. Therefore the complete set of free parameters is given by
the analysis dictionaries $\{ \bm{\Omega}_i \in \mathbb{R}^{d_i \times n_i} \}_{i=1}^L$, the soft-thresholds $\{ \bm{\lambda}_i \in \mathbb{R}^{d_i} \}_{i=1}^L$ and the synthesis dictionary $\bm{D} \in \mathbb{R}^{d_{L+1} \times  n_{L+1}}$
where $d_{L+1}= s^2$.

Fig. \ref{fig:overall} shows an example of a 2-layer DeepCAM for $2 \times$ image super-resolution. The input LR image denoted with $\bm{X}_{0}$ passes through multiple layers of convolution with the analysis dictionary and soft-thresholding. There are 4 synthesised HR sub-images $\widehat{\bm{Y}}$ which are obtained by convolving the last layer analysis feature maps with the synthesis dictionary $\bm{D}$ and will be rearranged to generate the final predicted HR image according to the sampling pattern.

Let us denote with $\bm{X}_{i-1} \in \mathbb{R}^{W_{i-1} \times H_{i-1} \times d_{i-1}}$ the input signal at the $i$-th layer, and denote with $\bm{\omega}_{i,j}^T \in \mathbb{R}^{p_i^2 d_{i-1}}$ the $j$-th atom of $\bm{\Omega}_i$. The convolution and soft-thresholding operations corresponding to the $j$-th atom and threshold pair $(\bm{\omega}_{i,j}, \bm{\lambda}_i(j))$ can be expressed as\footnote{This is a 2-D convolution performed along the image coordinate $H_{i-1}$, $W_{i-1}$.}:
\begin{equation}
    \begin{split}
        &\bm{F}_{i,j} = \text{mat}(\bm{\omega}_{i,j}) * \bm{X}_{i-1},\\
        &\bm{X}_{i,j} = \mathcal{S}_{\bm{\lambda}_i (j)}(\bm{F}_{i,j}),
    \end{split}
    \label{eq:convThre}
\end{equation}
where $\text{mat}(\cdot)$ represents the operation which reshapes a vector of length $p_i^2d_{i-1}$ to a tensor with size $p_i \times p_i \times d_{i-1}$, $\bm{F}_{i,j} \in \mathbb{R}^{W_i \times H_i}$ is the convolution result, $\mathcal{S}_{\lambda}(\cdot)$ denotes the element-wise soft-thresholding operation with threshold $\lambda$, and $\bm{X}_{i,j} \in \mathbb{R}^{W_i \times H_i}$ is the sparse representation after thresholding.

\begin{figure}
    \centering
    \includegraphics[width=0.45\textwidth]{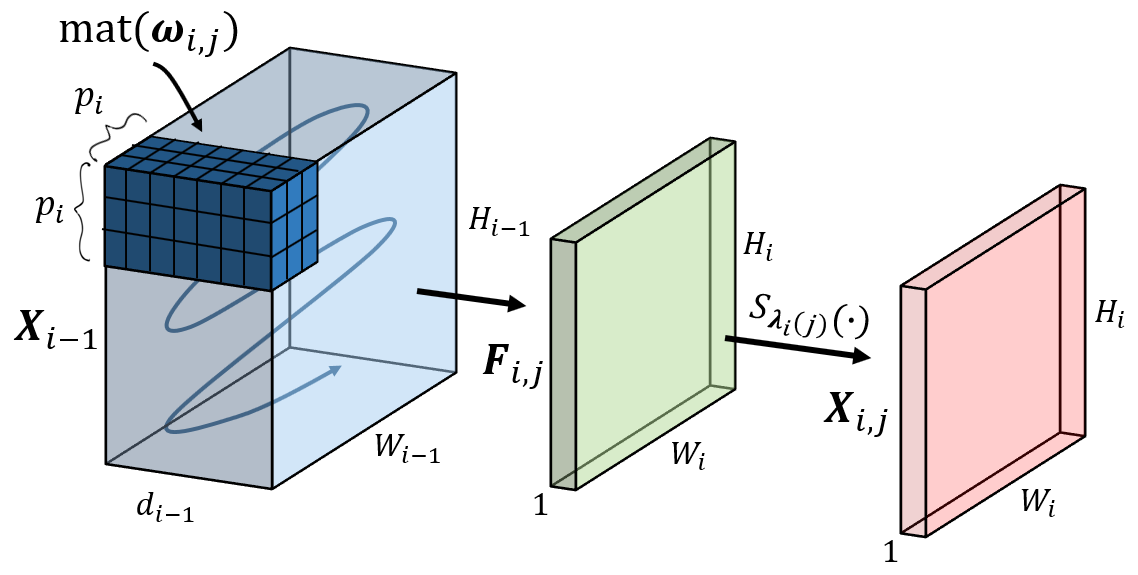}
    \caption{The convolution and soft-thresholding operation corresponding to the atom and threshold pair $(\bm{\omega}_{i,j}, \bm{\lambda}_i(j))$. The input signal $\bm{X}_{i-1}$ is of size $H_{i-1} \times W_{i-1} \times d_{i-1}$. The atom $\bm{\omega}_{i,j} \in \mathbb{R}^{p_i^2 d_{i-1}}$ represents a convolutional filter with spatial support size $p_i \times p_i$ and $d_{i-1}$ channels. The convolution of $\bm{\omega}_{i,j}$ and $\bm{X}_{i-1}$ results in a matrix $\bm{F}_{i,j}$ of size $W_i \times H_i$. An element-wise soft-thresholding operation $\mathcal{S}_{\bm{\lambda}_i (j)}(\cdot)$ is applied to every element of $\bm{F}_{i,j}$ and results in $\bm{X}_{i,j}$.}
    \label{fig:ConvFilter}
\end{figure}

Fig. \ref{fig:ConvFilter} illustrates the convolution and the soft-thresholding operation described in Eqn. (\ref{eq:convThre}). The convolution linearly transforms the input signal $\bm{X}_{i-1}$ to a 2-D representation $\bm{F}_{i,j}$. An element-wise soft-thresholding operation is then applied to every element on $\bm{F}_{i,j}$ and generates a sparse 2-D representation $\bm{X}_{i,j}$. 

By stacking the $d_{i}$ sparse 2-D representation $\{ \bm{X}_{i,j} \}_{j=1}^{d_i}$, the $i$-th layer output signal can be represented as $\bm{X}_i \in \mathbb{R}^{W_i \times H_i \times d_i}$. For simplicity, let us denote the $i$-th layer convolution and soft-thresholding operation as:
\begin{equation}
    \bm{X}_{i} = \mathcal{S}(\bm{\Omega}_i * \bm{X}_{i-1}, \bm{\lambda}_{i}).
    \label{eq:ConvSoft}
\end{equation}

When the convolution of $\bm{\Omega}_i$ and $\bm{X}_{i-1}$ is represented by a convolutional analysis dictionary $\bm{\mathcal{H}}(\bm{\Omega}_i, l)$ with $l =W_{i-1} H_{i-1} d_{i-1}$, the convolution and soft-thresholding operations can be expressed as follows:
\begin{equation}
    \bm{X}_{i} = \text{mat} \left( \mathcal{S}_{\bm{\lambda}_{i} \otimes \bm{1}}\left( \bm{\mathcal{H}}(\bm{\Omega}_i, l) \text{vec}(\bm{X}_{i-1}) \right) \right),
\end{equation}
where $\bm{1}$ is a all ones vector of size $W_{i}H_{i}$, and $\otimes$ is the Kronecker product.

The complete model of a $L$-layer DeepCAM can then be expressed as:
\begin{equation}
    \widehat{\bm{Y}} = \bm{D} * \mathcal{S} \left( \left( \cdots \mathcal{S}\left( \bm{\Omega}_1 * \bm{X}_{0}, \bm{\lambda}_{1}  \right) \cdots \right) ,\bm{\lambda}_{L} \right), \label{ddm}
\end{equation}
where $\widehat{\bm{Y}}$ denotes the estimated $d_{L+1}$ HR images.


\section{Learning A Deep Convolutional Analysis Dictionary Model}
In this section, we will introduce the proposed algorithm for learning both the convolutional analysis dictionary and the soft-thresholds in DeepCAM.

We adopt a joint \textit{Information Preserving} and \textit{Clustering} strategy 
as proposed in DeepAM \cite{huang2019DeepAM}. 
At each layer, the analysis dictionary $\bm{\Omega}_i$ is divided into two sub-dictionaries: an Information Preserving Analysis Dictionary (IPAD) $\bm{\Omega}_{\text{I}i}$ and a Clustering Analysis Dictionary (CAD) $\bm{\Omega}_{\text{C}i}$. The IPAD and soft-threshold pair $( \bm{\Omega}_{\text{I}i}, \bm{\lambda}_{\text{I}i})$ will generate feature maps that can preserve the information from the input image. The CAD and soft-threshold pair $( \bm{\Omega}_{\text{C}i}, \bm{\lambda}_{\text{C}i})$ will generate feature maps with strong discriminative power that can facilitate the prediction of the HR image. To achieve this goal, there should be a sufficient number of IPAD and CAD atoms and guidelines on how to determine the size of each dictionary will also be provided in this section.

\subsection{Learning IPAD and Threshold Pair}
\label{IPADLearn}

The Information Preserving Analysis Dictionary (IPAD) will be learned using the proposed convolutional analysis dictionary learning method of Section \ref{learnConvDict}. The thresholds will be set according to the method used in DeepAM \cite{huang2019DeepAM}.

A multi-layer convolutional analysis dictionary naturally possesses a multi-scale property.
The product of two convolutional dictionaries leads to a convolutional dictionary whose equivalent filters are given by the convolution of the filters in the two dictionaries due to the associativity property of convolution (i.e. $\bm{a} * ( \bm{b} * \bm{c} ) = (\bm{a} * \bm{b}) * \bm{c}$).

Let us denote with $\bm{\mathcal{H}}_i$ the $i$-th layer convolutional analysis dictionary constructed using convolutional filters with patch size $p_i$. 
The effective convolutional analysis dictionary $\bm{\mathcal{H}}^{(i)}=\bm{\mathcal{H}}_i \bm{\mathcal{H}}_{i-1} \cdots \bm{\mathcal{H}}_1$ has filters with {spatial patch} size:
\begin{equation}
    p_{\text{eff},i} = \sum_{j=1}^i p_j - (i-1).
    \label{eq:EffPatchSize}
\end{equation}

\begin{figure}[t]
    \centering
    \includegraphics[width=0.45\textwidth]{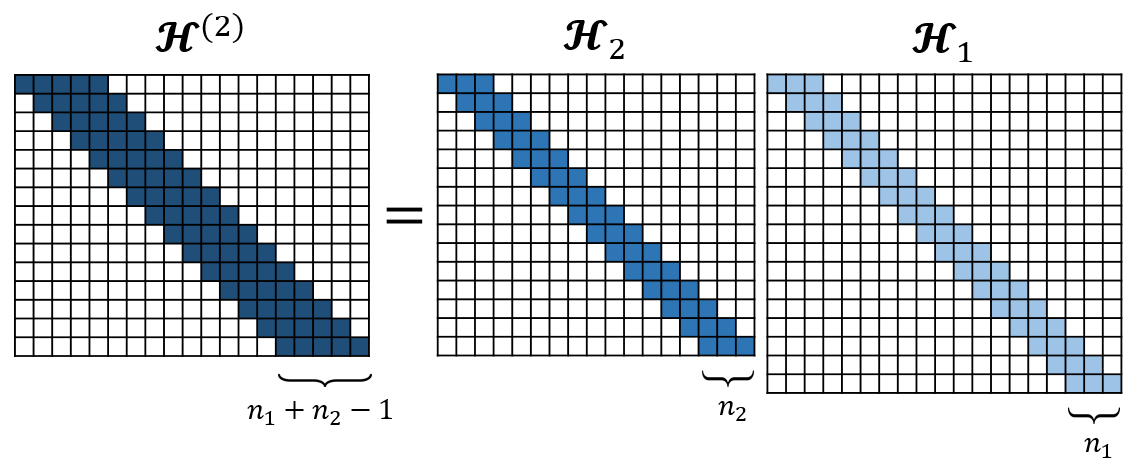}
    \caption{With a two-layer convolutional analysis dictionary, the effective convolutional analysis dictionary $\bm{\mathcal{H}}^{(2)}=\bm{\mathcal{H}}_2 \bm{\mathcal{H}}_1$ is still a convolutional analysis dictionary and with support size $n_2 + n_1 - 1$.}
    \label{fig:EffConvDict}
\end{figure}

An example of a two-layer convolutional analysis dictionary is shown in Fig. \ref{fig:EffConvDict}.
The effective dictionary $\bm{\mathcal{H}}^{(i)}$ has an effective patch size that increases with the number of layers and can be large even when each convolutional analysis dictionary uses filters with small patch size.

When the support size of a convolutional analysis dictionary is small, its row atoms can only receive local information from the whole input signal.
With an increased effective patch size, the row atoms of the convolutional analysis dictionary at a deeper layer will receive information from a larger segment of the input signal. 

For each HR pixel at the synthesised HR images $\widehat{\bm{Y}}$, there is a corresponding super-patch region on each layer which contributes all the information for predicting that pixel. Let us denote with $p_{S,i}$ the super-patch size at the $i$-th layer. It can be expressed in terms of the patch size of the convolutional filters from the final layer $L$ to layer $i$:
\begin{equation}
    p_{S,i} = p_{L+1} + \sum_{j=i}^{L} (p_j - 1).
    \label{eq:superpatchsize}
\end{equation}

\begin{figure}[t]
    \centering
    \includegraphics[width=0.3\textwidth]{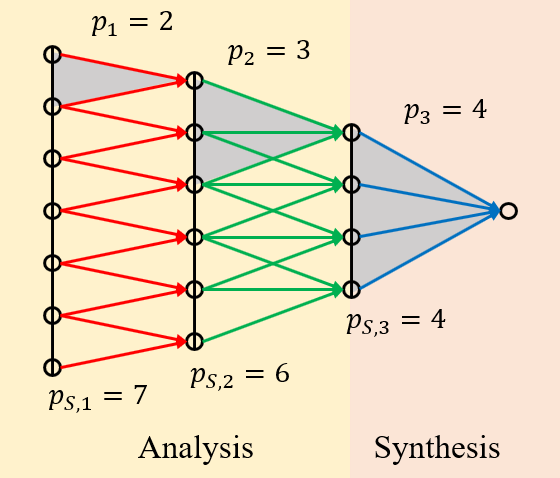}
    \caption{Super-patches at different layers in a $2$-layer DeepCAM. A synthesised pixel value corresponds to a super-patch on the $i$-th layer with patch size $p_{S,i}$. In this example, the convolutional filters are with size $p_1=2$, $p_2=3$, and $p_3 = 4$. The super-patch size at layer 1, 2, and 3 is 7, 6, and 4, respectively.}
    \label{fig:SuperPatch}
\end{figure}

Fig. \ref{fig:SuperPatch} shows the super-patches at different layers for a $2$-layer DeepCAM. Note that the super-patch size at a shallower layer is larger than that in a deeper layer.

In the proposed convolutional analysis dictionary learning method ConvGOAL+, there are two sets of training data: the super-patch training data $\bm{\mathcal{X}}_S$ and the small patch training data $\bm{\mathcal{X}}$. The super-patch data $\bm{\mathcal{X}}_S$ is used to impose the rank constraint. The small patch data $\bm{\mathcal{X}}$ has the same support as the filters and is used to impose the sparsifying and linear independence constraints.

The patch size of the super-patch training data for convolutional analysis dictionary learning should be no smaller than $p_{S,i}$. Otherwise, we can not ensure that the learned convolutional analysis dictionary will be able to utilize all information within the super-patch for predicting the corresponding HR pixel values.


At the $i$-th layer, let us define the support size of the super-patch training data as $S_i = p_{S,i}^2 d_{i-1}$.
The super-patch training data, the small patch training dataset and the ground-truth training dataset are denoted as $\bm{\mathcal{X}}_{S}^{i-1} \in \mathbb{R}^{S_i \times N_1}$, $\bm{\mathcal{X}}^{i-1} \in \mathbb{R}^{n_i \times N_2}$ and $\bm{\mathcal{Y}} \in \mathbb{R}^{s^2 \times N_2}$, respectively. 

Let us denote $d_{\text{I}i}$ as the number of atoms in $\bm{\Omega}_{\text{I}i}$. With $\bm{\Omega}_{\text{I}i} \in \mathbb{R}^{d_{\text{I}i} \times n_i}$, we will have $\bm{\mathcal{H}}(\bm{\Omega}_{\text{I}i}, S_i)\in 
\mathbb{R}^{d_{\text{I}i}(p_{S,i} - p_i + 1)^2 \times S_i}$. 
From the degree of freedom perspective, there should be at least $S_1 = p_{S,1}^2 d_{0}$ rows in $\bm{\mathcal{H}}(\bm{\Omega}_{\text{I}i}, S_i)$ to ensure that information from $\bm{\mathcal{X}}_{S}^0$ will be preserved. This leads to:
\begin{equation}
    d_{\text{I}i} \geq \frac{p_{S,1}^2 d_{0}}{(p_{S,i} - p_i + 1)^2}.
    \label{eq:miniAtoms}
\end{equation}

Eqn. (\ref{eq:miniAtoms}) indicates that there should be more atoms for information preservation in a deeper layer of a DeepCAM. For example, when $d_{0}=1$, $p_1=2$, $p_2=3$, and $p_3 = 4$ in a $2$-layer DeepCAM, there should be at least 2 atoms in the $1$-st layer, and 4 atoms in the $2$-nd layer for information preservation.

Given $d_{\text{I}i}$ atoms, the super-patch training data $\bm{\mathcal{X}}_{S}^{i-1}$ and the small patch training data $\bm{\mathcal{X}}^{i-1}$, the IPAD $\bm{\Omega}_{\text{I}i}$ is learned using ConvGOAL+ algorithm. 
The convolutional analysis dictionary $\bm{\mathcal{H}}(\bm{\Omega}_{\text{I}i},S_i)$ will then be able to preserve essential information from the input LR image.

The soft-thresholds $\bm{\lambda}_{\text{I}i}$ should be set properly. As in \cite{huang2019DeepAM}, the inner product between an analysis atom $\bm{\omega}_{i,j}$ and the small patch training samples $\bm{\mathcal{X}}^{i-1}$ can be well modelled by a Laplacian distribution with variance $\sigma_j$.
Therefore, as in \cite{huang2019DeepAM}, the soft-thresholds associated with IPAD $\bm{\Omega}_{\text{I}i}$ is set to be inversely proportional to the variances:
\begin{equation}
    \bm{\lambda}_{\text{I}i}=\rho_{\text{I}}\left[\frac{1}{\sigma_{1}},\frac{1}{\sigma_{2}},\cdots,\frac{1}{\sigma_{d_{\text{I}i}}}\right]^{T},
\end{equation}
where $\rho_{\text{I}}$ is a scaling parameter, and the variance $\sigma_{j}$ of the $j$-th coefficient can be estimated using the obtained IPAD $\bm{\Omega}_{\text{I}i}$ and the small patch training data $\bm{\mathcal{X}}^{i-1}$. 

The free parameter $\rho_{\text{I}}$ is determined by solving a 1-dimensional search problem. The optimization problem for $\rho_{\text{I}}$ is therefore formulated as:
\begin{equation}
    {\rho_{\text{I}}}=\arg\underset{\rho \in \mathcal{D}}{\min}\left\Vert \bm{\mathcal{Y}}-\bm{G}\mathcal{S}_{\rho \bm{\lambda} \otimes \bm{1}}\left( \bm{\mathcal{H}}(\bm{\Omega}_{\text{I}i},S_i) \text{vec}(\bm{\mathcal{X}}_S^{i-1}) \right)\right\Vert _{F}^{2},
    \label{eq:1D-opt_DeepCAMIPAD}
\end{equation}
where $\bm{\lambda}=\left[{1}/{\sigma_{1}},{1}/{\sigma_{2}},\cdots,{1}/{\sigma_{d_{\text{I}i}}}\right]^T$, $\bm{1}$ is an all ones vector of size $(p_{S,i} - p_i + 1)^2$, $\otimes$ is the Kronecker product, $\bm{G}=\bm{\mathcal{Y}}\bm{Z}^{T}(\bm{Z}\bm{Z}^{T})^{-1}$ with $\bm{Z}=\mathcal{S}_{\rho\bm{\lambda} \otimes \bm{1}}\left( \bm{\mathcal{H}}(\bm{\Omega}_{\text{I}i},S_i) \text{vec}(\bm{\mathcal{X}}_S^{i-1}) \right)$, and $\mathcal{D}$ is a discrete set of values. 


\subsection{Learning CAD and Threshold Pair}

The objective of a Clustering Analysis Dictionary (CAD) is to perform a linear transformation to its input signal such that the responses are highly correlated with the most significant components of the residual error. Soft-thresholding, which is used as the non-linearity, sets to zero the data with relatively small responses. The components with large residual error will then be identified.

The number of atoms in $\bm{\Omega}_{\text{C}i}$ is essential to the performance of DeepCAM. Similar to the discussions in Section \ref{IPADLearn}, with $d_{\text{C}i}$ atoms in $\bm{\Omega}_{\text{C}i}$, the size of the convolutional analysis dictionary $\bm{\mathcal{H}}(\bm{\Omega}_{\text{C}i}, S_i)$ will be $d_{\text{C}i}(p_{S,i} - p_i + 1)^2 \times S_i$. 
For each super-patch region on the LR image, the number of coefficients for discriminative feature representation should not decrease over layers. That is, we would like to have more atoms in $\bm{\mathcal{H}}(\bm{\Omega}_{\text{C}i},S_i)$ than in $\bm{\mathcal{H}}(\bm{\Omega}_{\text{C}i-1},S_{i-1})$. Therefore, the number of CAD atoms should meet the condition:
\begin{equation}
    d_{\text{C}i} \geq d_{\text{C}i-1} \frac{(p_{S,i-1} - p_{i-1} + 1)^2}{(p_{S,i} - p_i + 1)^2}.
    \label{eq:miniAtomsCAD}
\end{equation}

Different from the unstructured deep dictionary model, it is not straightforward to set the dictionary sizes.
Eqn. (\ref{eq:miniAtoms}) and Eqn. (\ref{eq:miniAtomsCAD}) provide a guideline on how to set the number of atoms in order to generate representations that are both information preserving and discriminative.


Let us denote with $\bm{\mathcal{Y}}^{i} \in \mathbb{R}^{s^2p_i^2 \times N_2}$ the corresponding HR patch training data of $\bm{\mathcal{X}}^{i-1}$.
A synthesis dictionary $\bm{D}_{i} \in \mathbb{R}^{s^2p_i^2 \times n_i}$ can be learned to map $\bm{\mathcal{X}}^{i-1}$ to $\bm{\mathcal{Y}}^{i}$ by solving:
\begin{equation}
    \bm{D}_{i} = \arg \underset{\bm{D}}{\min} ||\bm{D}\bm{\mathcal{X}}^{i-1}-\bm{\mathcal{Y}}^{i}||^2_F. 
\end{equation}
It has a closed-form solution: 
\begin{equation}
    \bm{D}_i=\bm{\mathcal{Y}}^{i}{\bm{\mathcal{X}}^{i-1}}^{T}\left(\bm{\mathcal{X}}^{i-1}{\bm{\mathcal{X}}^{i-1}}^{T}\right)^{-1}.\label{eq:synthesis_i}
\end{equation}

Given $\bm{D}_{i}$, we define the middle resolution (MR) and the residual data as $\widehat{{\bm{\mathcal{Y}}}}^{i} = \bm{D}_{i}\bm{\mathcal{X}}^{i-1}$ and $\bm{\mathcal{E}}^{i}=\bm{\mathcal{Y}}^{i} - \widehat{{\bm{\mathcal{Y}}}}^{i}$, respectively.
The MR data is a linear transformation of the input small patch training data. The residual data contains the information about the residual energy.

We propose to learn an analysis dictionary $\bm{\Psi}_{i} \in \mathbb{R}^{d_{\text{C}i} \times s^2p_i^2}$ in the ground-truth data domain. If $\bm{\Psi}_{i}$ is able to simultaneously sparsify the middle resolution data $\widehat{{\bm{\mathcal{Y}}}}^{i}$ and the residual data $\bm{\mathcal{E}}^{i}$, the atoms within the learned $\bm{\Psi}_i$ will then be able to identify the data in $\widehat{{\bm{\mathcal{Y}}}}^{i}$ with large residual energy and the $i$-th layer CAD is then re-parameterized as:
\begin{equation}
    \label{eq:repara}
    \bm{\Omega}_{\text{C}i} = \bm{\Psi}_i\bm{D}_{i}.
\end{equation}

Therefore an additional constraint as proposed in \cite{huang2019DeepAM} is applied to impose the simultaneous sparsifying property.
Each analysis atom $\bm{\psi}_{k}^T$ is enforced to be able to jointly sparsify $\widehat{{\bm{\mathcal{Y}}}}^{i}$ and $\bm{\mathcal{E}}^{i}$:
\begin{equation}\label{eq:sparsify2}
    p(\bm{\Psi}) = c\sum_{j=1}^{N_2} \sum_{k=1}^{d_{\text{C}i}}\log \left(1 + \nu \left(\left(\bm{\psi}_{k}^T \widehat{\bm{y}}^{i}_{j})^2 - (\bm{\psi}_{k}^T \bm{e}^{i}_{j}\right)^2\right)^2 \right),
\end{equation}
where $c=1/{N_2d_{\text{C}i} \log(1+\nu)}$, $\nu$ is a tunable parameter, and $\widehat{\bm{y}}^{i}_{j}$ and $\bm{e}^{i}_{j}$ are the $j$-th column of $\widehat{{\bm{\mathcal{Y}}}}^{i}$ and $\bm{\mathcal{E}}^{i}$, respectively.

The objective function for learning the analysis dictionary $\bm{\Psi}_{i}$ can then be formulated as:
\begin{equation}
    \bm{\Psi}_{i}=\arg\underset{\bm{\Psi}^{T}\in \Theta_{{\Psi}}}{\min}f(\bm{\Psi}),\label{eq:GOAL+2}
\end{equation}
where $f(\bm{\Psi})=g(\bm{\Psi})+\kappa h(\bm{\Psi})+\upsilon l(\bm{\Psi})  + \mu p(\bm{\Psi})$ with $\kappa$, $\upsilon$ and $\mu$ being the regularization parameters. The functions $g(\bm{\cdot})$, $h(\bm{\cdot})$ and $l(\bm{\cdot})$ are those defined in Eqn. (\ref{eq:sparsify}), Eqn. (\ref{eq:log-det2}) and Eqn. (\ref{eq:log-barrier}), respectively.
To have zero mean responses for each learned CAD atom, the feasible set of the analysis dictionary $\bm{\Psi}$ is set to $\Theta_{{\Psi}}=\mathbb{S}_{d_{\text{C}i}-1}^{s^2p_i^2}\cap\bm{V}^{\perp}$ with $\bm{V} = \bm{1} \in \mathbb{R}^{s^2p_i^2}$.

The objective function in Eqn. (\ref{eq:GOAL+2}) is optimized using ConvGOAL+ algorithm. With the learned analysis dictionary $\bm{\Psi}_{i}$, the $i$-th layer CAD is then obtained as in Eqn. (\ref{eq:repara}).

As proposed in DeepAM \cite{huang2019DeepAM}, it is both effective and efficient to set CAD soft-thresholds being proportional to the variance of the analysis coefficients. The CAD soft-thresholds are therefore defined as follows:
\begin{equation}
    \bm{\lambda}_{\text{C}i} = \rho_{\text{C}} \left[{\sigma_{1}}, {\sigma_{2}}, \cdots, {\sigma_{d_{\text{C}i}}} \right],\label{eq:para2}
\end{equation}
where $\rho_{\text{C}}$ is a scaling parameter, and $\sigma_{j}$ is the variance of the Laplacian distribution for the $j$-th atom.

The free parameter $\rho_{\text{C}}$ can be learned using a similar approach to the one used to solve Eqn. (\ref{eq:1D-opt_DeepCAMIPAD}). 
As the analysis coefficients can be well modelled by Laplacian distributions, the proportion of data that is set to zero for each pair of atom and threshold will be the same.
The optimization problem for $\rho_{\text{C}}$ is formulated as:
\begin{equation}
    {\rho_{\text{C}}}=\arg\underset{\rho \in \mathcal{D}}{\min}\left\Vert {\bm{\mathcal{Y}}}_{R}-\bm{G}\mathcal{S}_{\rho \bm{\lambda} \otimes \bm{1}}\left( \bm{\mathcal{H}}(\bm{\Omega}_{\text{C}i},S_i) \text{vec}(\bm{\mathcal{X}}_S^{i-1}) \right)\right\Vert _{F}^{2},
    \label{eq:1D-opt_DeepCAMCAD}
\end{equation}
where ${\bm{\mathcal{Y}}}_{R}$ is the estimation residual using IPAD, $\bm{\lambda}=\left[{\sigma_{1}},{\sigma_{2}},\cdots,{\sigma_{d_{\text{C}i}}}\right]^T$, $\bm{1}$ is a all ones vector of size $(p_{S,i} - p_i + 1)^2$, $\otimes$ is the Kronecker product, $\bm{G}=\bm{\mathcal{Y}}_{R}\bm{Z}^{T}(\bm{Z}\bm{Z}^{T})^{-1}$ with $\bm{Z}=\mathcal{S}_{\rho \bm{\lambda} \otimes \bm{1}}\left( \bm{\mathcal{H}}(\bm{\Omega}_{\text{C}i},S_i) \text{vec}(\bm{\mathcal{X}}_S^{i-1}) \right)$, and $\mathcal{D}$ is a discrete set of values.

\subsection{Synthesis Dictionary Learning}

At the last layer, the synthesis dictionary $\bm{D} \in \mathbb{R}^{s^2 \times n_{L+1}}$ will transform the $L$-th layer deep convolutional representation $\bm{\mathcal{X}}^{L} \in \mathbb{R}^{n_{L+1} \times N_2}$ to the ground-truth training data $\bm{\mathcal{Y}} \in \mathbb{R}^{s^2 \times N_2}$. The synthesis dictionary can be learned using least squares:
\begin{equation}
    \bm{D}=\bm{\mathcal{Y}}{\bm{\mathcal{X}}^{L}}^{T}\left(\bm{\mathcal{X}}^{L}{\bm{\mathcal{X}}^{L}}^{T}\right)^{-1}.\label{eq:synthesis}
\end{equation}

Convolving the learned synthesis dictionary with the $L$-th layer feature maps, results in $s^2$ estimated HR images which can be reshaped and combined to form the final estimated HR image.


The overall learning algorithm for DeepCAM is summarized in \textbf{Algorithm \ref{DeepCAMAlg}}. 

\begin{algorithm}[t]\label{DeepCAMAlg}
    \SetAlgoLined
    \textbf{Input:} Training data pair $(\bm{\mathcal{X}}^{0},\bm{\mathcal{Y}})$, the number of layers $L$, and the structure of DeepCAM;
    
    \For{$i\gets 1$ \KwTo $L$}{
        Learning $\bm{\Omega}_{\text{I}i}$ using ConvGOAL+ with training data $(\bm{\mathcal{X}}_S^{i-1},\bm{\mathcal{X}}^{i-1})$ and objective function Eqn. (\ref{eq:GOAL+}); 

        Learning $\bm{\Omega}_{\text{C}i}$ using ConvGOAL+ with training data $(\bm{\mathcal{X}}_S^{i-1},\bm{\mathcal{X}}^{i-1},\bm{\mathcal{Y}})$ and objective function Eqn. (\ref{eq:GOAL+2});  
        
        Learning thresholds $\bm{\lambda}_{\text{I}i}$ first and then $\bm{\lambda}_{\text{C}i}$;
        
        $\bm{\Omega}_i\gets [\bm{\Omega}_{\text{I}i};\bm{\Omega}_{\text{C}i}]$, $\bm{\lambda}_i\gets [\bm{\lambda}_{\text{I}i};\bm{\lambda}_{\text{C}i}]$;
        
        Update the super-patch training data as $\bm{\mathcal{X}}_S^{i} = \mathcal{S}(\bm{\Omega}_i * \bm{\mathcal{X}}_S^{i-1}, \bm{\lambda}_{i})$;
        
        Extract the small patch training data $\bm{\mathcal{X}}^{i}$ from $\bm{\mathcal{X}}_S^{i}$;
    }
    Learning the synthesis dictionary $\bm{D}$ as in Eqn. (\ref{eq:synthesis}).

    \textbf{Output:} Learned DeepCAM $\left\{ \left\{ \bm{\Omega}_{i}, \bm{\lambda}_{i}\right\} _{i=1}^{L}, \bm{D} \right\}$.
 \caption{DeepCAM Learning Algorithm}
\end{algorithm}

\section{Simulation Results}

In this section, we report the implementation details and numerical results of our proposed DeepCAM method and compare it with other existing single image super-resolution algorithms.

\subsection{Implementation Details}

Most of the implementation settings are the same as in \cite{huang2019DeepAM}.
The standard 91 training images \cite{yang2010image} are used as the training dataset and the \textit{Set5} \cite{yang2010image} and the \textit{Set14} \cite{zeyde2010single} are used as the testing datasets. The color images have been converted from the RGB color space to the YCbCr color space. Image super-resolution is only performed on the luminance channel. 

Table \ref{paramSet} shows the parameters setting of ConvGOAL+ algorithm for learning the $i$-th layer IPAD and CAD. 
Both the IPAD and the CAD are initialized with i.i.d. Gaussian random entries.
The spatial size of the super-patches $p_{S,i}$ used for training is set to the minimum value as indicated by Eqn. (\ref{eq:superpatchsize}) for the purpose of minimizing the training computational complexity. The number of IPAD atoms $d_{\text{I}i}$ is set to the minimum integer satisfying Eqn. (\ref{eq:miniAtoms}). The number of CAD atoms is then set to $d_{\text{C}i} = d_{i} - d_{\text{I}i}$ with a pre-defined total number of channel $d_i$.
We apply batch training for ConvGOAL+ algorithm. The training data has been equally divided into $N_b = 10$ batches. During training, the ConvGOAL+ algorithm is sequentially applied to each batch until the learned dictionary converges or all batches have been used for training. 
For each batch, $\tau = 100$ iterations of conjugate gradient descent is performed to update the dictionary. The discrete set $\mathcal{D}$ used for searching the scaling parameter of the thresholds is set to be $\mathcal{D} = [\cdots,10^{-2},2\times10^{-2},\cdots,10^{-1},2\times10^{-1},\cdots]$.

\begin{table}
    \center
    \begin{tabular}{|c| c|c|c|c|}
    \hline 
    {Parameters} & {$\nu$} & {$\kappa$} & {$\upsilon$} & {$\mu$}\tabularnewline
    \hline \hline
    {IPAD} & {$10\times n_{i}$} & {$100 \times d_{\text{I}i}$} & {$0.01\times d_{\text{I}i}$} & {---}\tabularnewline
    \hline
    {CAD} & {$10\times n_{i}$} & {$0.1\times d_{\text{C}i}$} & {$0.01\times d_{\text{C}i}$} & {100}\tabularnewline
    \hline 
    \end{tabular}{\par}

    \caption{Parameters setting of GOAL+ algorithm for learning the $i$-th layer IPAD $\bm{\Omega}_{\text{I}i} \in \mathbb{R}^{d_{\text{I}i} \times n_{i}}$ and CAD $\bm{\Omega}_{\text{C}i} \in \mathbb{R}^{d_{\text{C}i} \times n_{i}}$. }
\label{paramSet}
\end{table}

\subsection{Analysis of the Learned DeepCAM}

In this section, we analyze the learned DeepCAM in terms of the number of layers, learned soft-thresholds and extracted feature maps. 

\begin{table}[]
    \centering
    \begin{tabular}{|c|c|c|c|}
\hline
Layer Number    & 1 & 2 & 3\\ \hline
Filter Number       & {[64]}       & {[16,64]}    & {[9,25,64]}  \\ \hline \hline
baby             & 38.03          & 38.21          & 38.30          \\ \hline
bird             & 39.74          & 40.19          & 40.60          \\ \hline
butterfly        & 31.70          & 31.99          & 32.17          \\ \hline
head             & 35.54          & 35.58          & 35.61          \\ \hline
woman            & 34.76          & 34.83          & 34.84          \\ \hline
\textbf{Average} & {35.95} & {36.16} & {36.30} \\ \hline
\end{tabular}
\caption{PSNR (dB) of DeepCAM with different number of layers. For all DeepCAM, the spatial filter size at all layers is $3 \times 3$. The maximum number of filters at the last layer is set to 64. $[d_1,d_2]$ denotes that there are $d_1$ filters at the first layer, and $d_2$ filter at the second layer.}
\label{tab:DeepCAM_layers}
\end{table}

\begin{figure*}[t]
    \centering
    \hspace*{\fill}
	\subfigure[Layer 1 soft-thresholds.]{
		\label{fig:T1} 
		\includegraphics[width=0.3\textwidth]{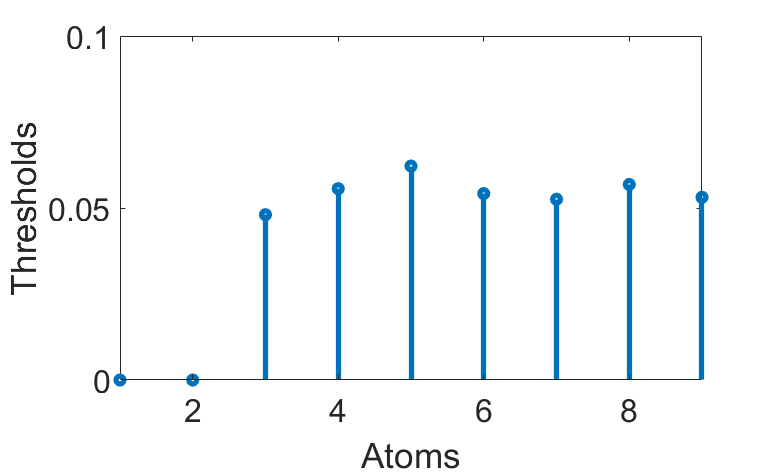}}
	\hfill
	\subfigure[Layer 2 soft-thresholds.]{
		\label{fig:T2} 
		\includegraphics[width=0.3\textwidth]{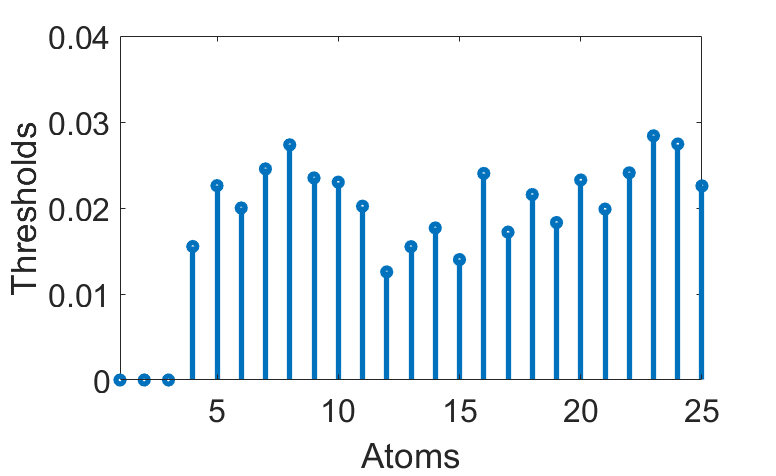}}
	\hfill
	\subfigure[Layer 3 soft-thresholds.]{
		\label{fig:T3} 
		\includegraphics[width=0.3\textwidth]{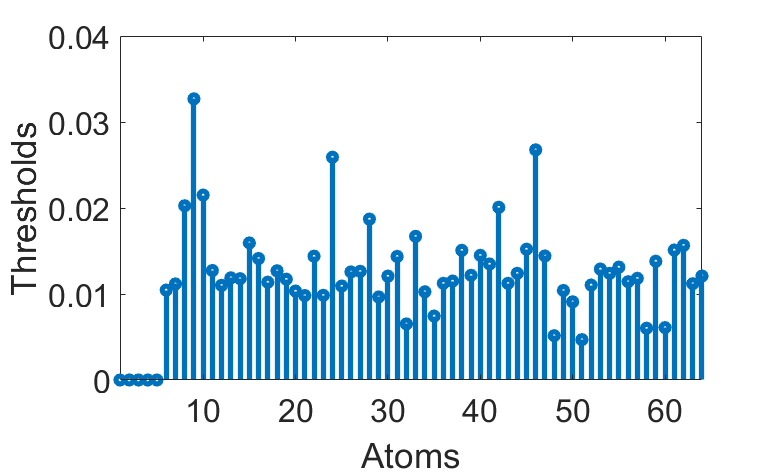}}
	\hspace*{\fill}
    \caption{{The soft-thresholds in layer 1, 2 and 3 of DeepCAM.} There is a bimodal behaviour on the thresholds. The thresholds corresponding to IPAD are relatively small, while the thresholds corresponding to CAD are relatively large.}
    \label{fig:ThresholdDeepCAM}
\end{figure*}

\begin{figure*}[t]
    \centering
    \hspace*{\fill}
	\subfigure[]{
		\label{fig:F1} 
		\includegraphics[width=0.23\textwidth]{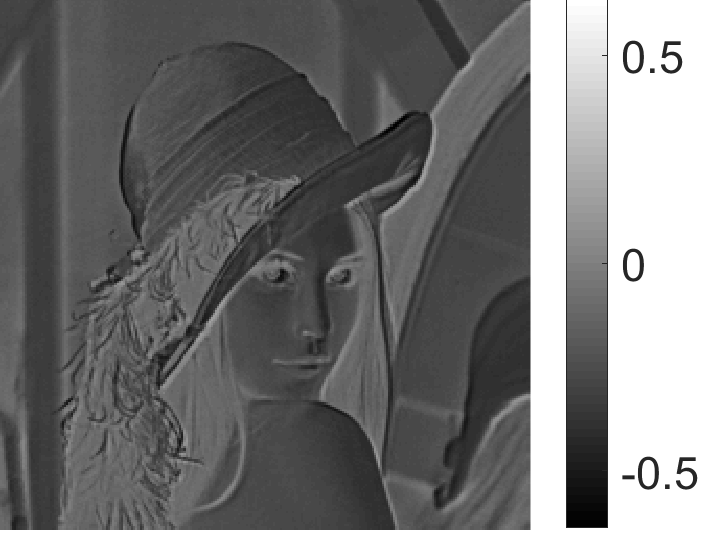}}
	\hfill
	\subfigure[]{
		\label{fig:F2} 
		\includegraphics[width=0.23\textwidth]{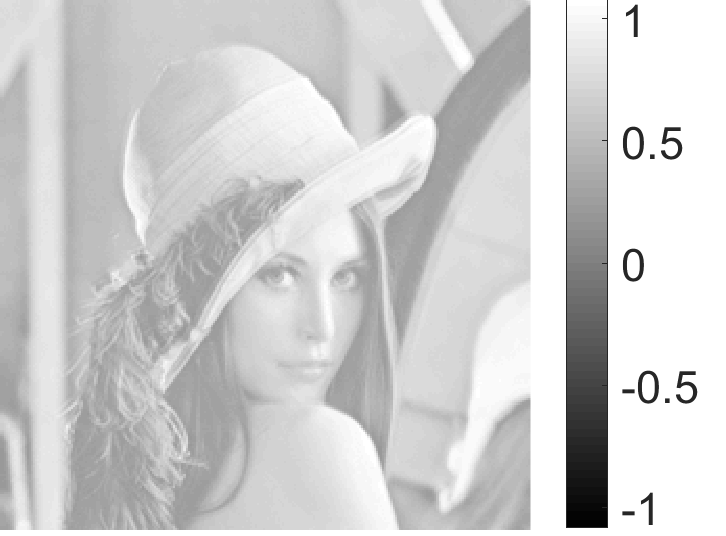}}
	\hfill
	\subfigure[]{
		\label{fig:F3} 
		\includegraphics[width=0.23\textwidth]{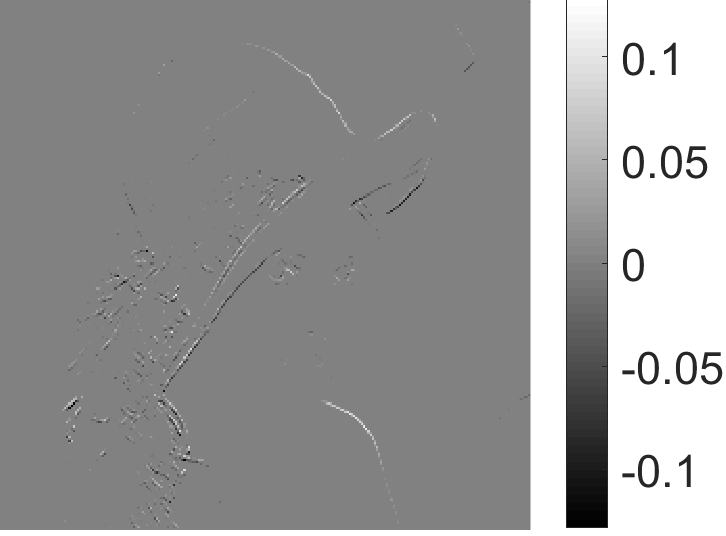}}
	\hfill
	\subfigure[]{
		\label{fig:F4} 
		\includegraphics[width=0.23\textwidth]{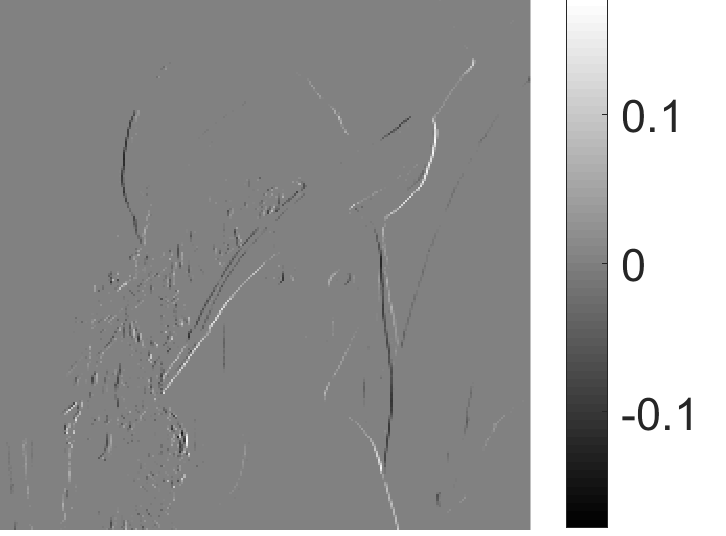}}
	\hspace*{\fill}
	
    \caption{{The first layer feature maps of DeepCAM.} The feature maps in \ref{fig:F1} - \ref{fig:F2} are due to IPAD and contain detailed structural information about the input LR image. The feature maps in \ref{fig:F3} - \ref{fig:F4} are due to CAD. They contain directional edges.}
    \label{fig:FeatureMaps}
\end{figure*}

Table \ref{tab:DeepCAM_layers} shows the PSNR (dB) of the learned DeepCAM with different number of layers evaluated on \textit{Set5} \cite{zeyde2010single}. For DeepCAM with different number of layers, the spatial filter size is set to $3 \times 3$ for all layers and the maximum number of filters at the last layer is set to 64. The effective filter size for the DeepCAM with 1, 2, and 3 layers is therefore $5 \times 5$, $7 \times 7$, and $9 \times 9$, respectively.
We can see that DeepCAM with more layers achieves higher average PSNR. From 1 layer to 2 layers, there is an improvement of about 0.2 dB. In particular, the PSNR of the testing image “bird” has been improved by around 0.5 dB. With 3 layers, further improvements can be observed on all testing images.
The improved performance of DeepCAM with more layers can be due to two reasons.
First, a deeper model has more non-linear layers and has therefore a stronger expressive power. 
Second, different from the unstructured deep dictionary model, a deeper convolutional dictionary model has an increased effective filter size which helps include more information for prediction and therefore improves prediction performance.

Fig. \ref{fig:ThresholdDeepCAM} shows the soft-thresholds of a 3-layer DeepCAM with 9, 25 and 64 filters at layer 1, 2 and 3, respectively. We can observe that the soft-thresholds have a bimodal behaviour. That is, the thresholds corresponding to IPAD are relatively small, while the thresholds corresponding to CAD are relatively large. 
Another observation is that the amplitude of the soft-thresholds decreases over layers. This will lead to denser representations at a deeper layer which can represent more complex signals.

Due to different learning objectives, the resultant feature maps of IPAD and CAD contains different information.
Fig. \ref{fig:FeatureMaps} shows the feature maps of the first layer in a 3-layer DeepCAM. The first 2 feature maps correspond to IPAD. We can find that these two feature maps, especially, the feature map in Fig. \ref{fig:F2} represents detailed structural information of the input LR image.
The feature maps in Fig.\ref{fig:F3} and \ref{fig:F4} are due to CAD and have zero responses on most regions due to relatively large soft-thresholds. These maps contain different directional edges corresponding to regions that require non-linear estimations. 
A combination of these features from both IPAD and CAD forms an informative and discriminative feature representation for predicting the HR image.

\begin{table*}[t]
    \centering
    \begin{tabular}{|c|c|c|c|c|c|c|}
    \hline
    Method               & {SC \cite{zeyde2010single}}    & {ANR \cite{timofte2013anchored}}       & {A+ \cite{timofte2014a+}}        & {SRCNN \cite{dong2014learning}} & DeepAM  & DeepCAM \\ \hline
    Parameters &     65,536 & 1,064,896 & 1,064,896 & 8,128 & 156,672 & 34,740  \\ \hline
    \end{tabular}
    \caption{Number of free parameters in different single image super-resolution methods.}
    \label{tab:NumParam}
\end{table*}

\begin{table*}[t]
    \centering
    \begin{tabular}{|c|c|c|c|c|c|c|c|c|c|c|c|}
    \hline 
    {Images} &  {Bicubic} &  {SC \cite{zeyde2010single}} &  {ANR \cite{timofte2013anchored}} &  {A+ \cite{timofte2014a+}} &  {SRCNN \cite{dong2014learning}} &  {DeepAM \cite{huang2019DeepAM}} & {$\text{CNN-BP}_{80}$} & {$\text{CNN-BP}_{40}$} & {$\text{DeepCAM}$} & {$\text{DeepCAM}_{\text{bp}}$} \tabularnewline
    \hline 
baby      & 36.93   & 38.11 & 38.31 & 38.39 & 38.16 & 38.48  & 38.31       & 38.31      & 38.18   & 38.20      \\
bird      & 36.76   & 39.87 & 40.00 & 41.11 & 40.58 & 41.01  & 41.10       & 41.06      & 40.68   & 40.78      \\
butterfly & 27.58   & 30.96 & 30.76 & 32.37 & 32.58 & 31.44  & 32.23       & 32.56      & 32.39   & 32.50      \\
head      & 34.71   & 35.47 & 35.54 & 35.64 & 35.51 & 35.64  & 35.57       & 35.59      & 35.50   & 35.55      \\
woman     & 32.31   & 34.59 & 34.68 & 35.44 & 35.07 & 35.20  & 34.84       & 35.16      & 34.93   & 35.25      \\
\hline 
\textbf{Average}   & 33.66   & 35.80 & 35.86 & 36.59 & 36.38 & 36.35  & 36.41       & 36.54      & 36.34   & 36.45    \\
\hline 

\end{tabular}
    \caption{PSNR (dB) of different methods evaluated on \textit{Set5} \cite{yang2010image}. 
    }
    \label{tab:Set5_DeepCAM}
\end{table*}

\begin{table*}[t]
    \centering
    \begin{tabular}{|c|c|c|c|c|c|c|c|c|c|c|c|}
    \hline 
    {Images} &  {Bicubic} &  {SC \cite{zeyde2010single}} &  {ANR \cite{timofte2013anchored}} &  {A+ \cite{timofte2014a+}} &  {SRCNN \cite{dong2014learning}} &  {DeepAM \cite{huang2019DeepAM}} & {$\text{CNN-BP}_{80}$} & {$\text{CNN-BP}_{40}$} & {$\text{DeepCAM}$} & {$\text{DeepCAM}_{\text{bp}}$} \tabularnewline
    \hline 
baboon      & 24.85 & 25.47 & 25.55 & 25.66 & 25.64 & 25.65 & 25.54 & 25.63 & 25.52 & 25.55      \\
barbara     & 27.87 & 28.50 & 28.43 & 28.49 & 28.53 & 28.49 & 27.88 & 28.14 & 28.45 & 28.40      \\
bridge      & 26.64 & 27.63 & 27.62 & 27.87 & 27.74 & 27.82 & 27.82 & 27.87 & 27.81 & 27.84      \\
c.guard  & 29.16 & 30.23 & 30.34 & 30.34 & 30.43 & 30.44 & 30.28 & 30.37 & 30.40 & 30.40      \\
comic       & 25.75 & 27.34 & 27.47 & 27.98 & 28.17 & 27.77 & 27.29 & 27.82 & 27.72 & 27.98      \\
face        & 34.69 & 35.45 & 35.52 & 35.63 & 35.57 & 35.62 & 35.54 & 35.55 & 35.48 & 35.51      \\
flowers     & 30.13 & 32.04 & 32.06 & 32.80 & 32.95 & 32.45 & 32.65 & 32.82 & 32.57 & 32.78      \\
foreman     & 35.55 & 38.41 & 38.31 & 39.45 & 37.43 & 38.89 & 39.31 & 39.24 & 39.12 & 39.23      \\
lenna       & 34.52 & 36.06 & 36.17 & 36.45 & 36.46 & 36.46 & 36.35 & 36.39 & 36.16 & 36.25      \\
man         & 29.11 & 30.31 & 30.33 & 30.74 & 30.78 & 30.57 & 30.68 & 30.71 & 30.58 & 30.70      \\
monarch     & 32.68 & 35.50 & 35.46 & 36.77 & 37.11 & 36.06 & 36.62 & 36.84 & 36.56 & 36.85      \\
pepper      & 35.02 & 36.64 & 36.51 & 37.08 & 36.89 & 36.87 & 36.99 & 36.99 & 36.86 & 36.97      \\
ppt3        & 26.58 & 29.00 & 28.67 & 29.79 & 30.31 & 29.13 & 28.02 & 29.18 & 29.61 & 29.67      \\
zebra       & 30.41 & 33.05 & 32.91 & 33.45 & 33.14 & 33.34 & 32.45 & 33.04 & 33.33 & 33.28      \\
\hline
\textbf{Average} & {30.21} & {31.83} & {31.81} & {32.32} & {32.22} & {32.11} & {31.96} & {32.18} & {32.16} & {32.24}     \\
\hline 

\end{tabular}
    \caption{PSNR (dB) of different methods evaluated on \textit{Set14} \cite{zeyde2010single}.
    }
    \label{tab:Set14_DeepCAM}
\end{table*}

\subsection{Comparison with Single Image Super-Resolution Methods}

In this section, we compare our proposed DeepCAM method with the DeepAM \cite{huang2019DeepAM} and some existing single image super-resolution methods including bicubic interpolation, sparse coding (SC)-based method \cite{zeyde2010single}, Anchored Neighbor Regression (ANR) \cite{timofte2013anchored}, Adjusted Anchored Neighborhood Regression (A+) \cite{timofte2014a+}, and Super-Resolution Convolutional Neural Network (SRCNN) \cite{dong2014learning}.

\begin{figure*}[t]
    \centering
    \hspace*{\fill}
	\subfigure[$\text{CNN-BP}_{80}$ (PSNR = 28.02 dB).]{
	    \label{fig:LR_DeepCAM}
		\includegraphics[width=0.32\textwidth]{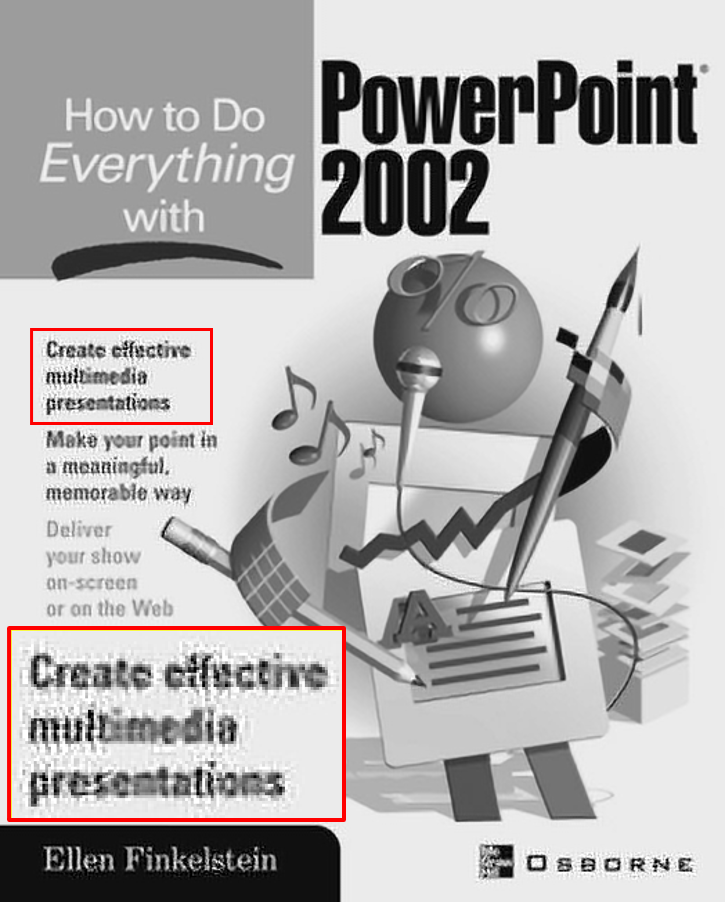}}
	\hfill
	\subfigure[$\text{CNN-BP}_{40}$ (PSNR = 29.18 dB).]{
	    \label{fig:DeepAM_DeepCAM}
		\includegraphics[width=0.32\textwidth]{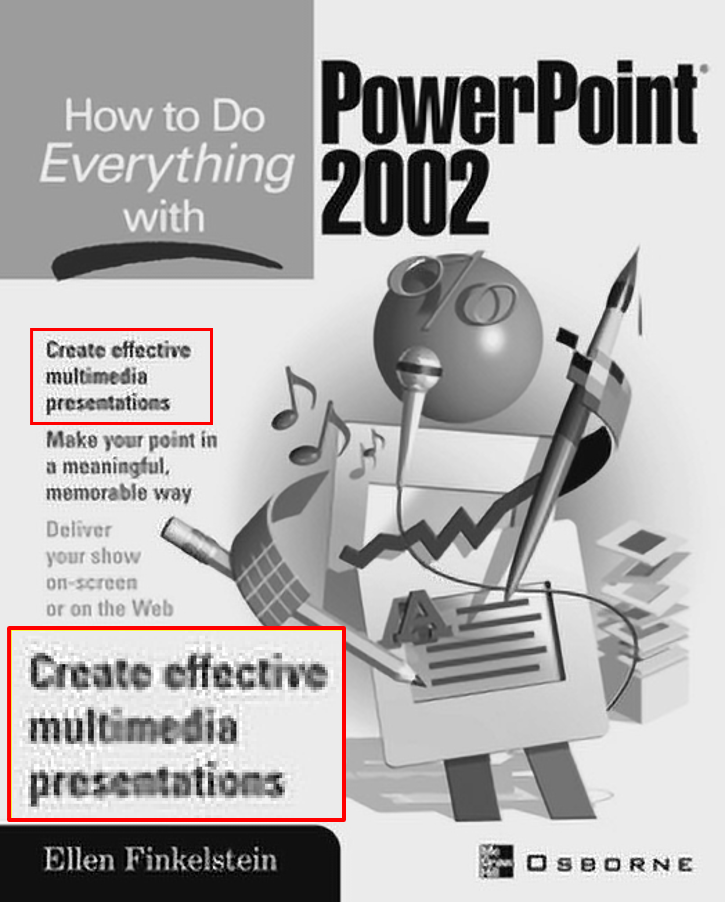}}
	\hfill
	\subfigure[DeepCAM (PSNR = 29.65 dB).]{
	    \label{fig:DeepCAM_DeepCAM}
		\includegraphics[width=0.32\textwidth]{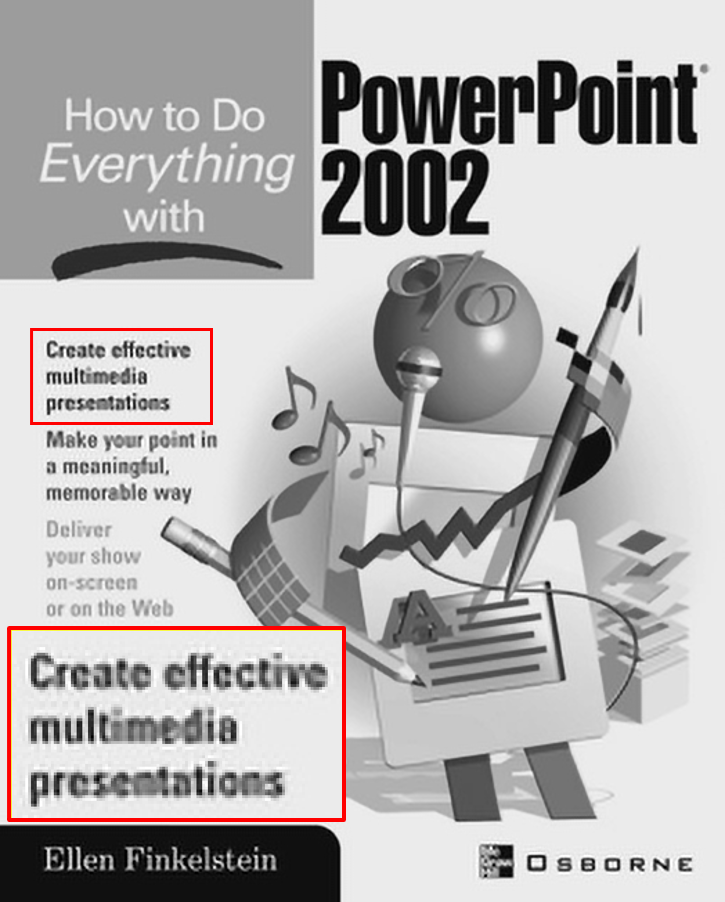}}
	\hspace*{\fill}
	
    \caption{{Examples of reconstructed HR images by different methods.} DeepCAM achieves better results than the backpropagation trained CNN. A region with characters on the reconstructed image have been marked using red rectangle and zoomed in.}
    \label{fig:DeepCAM_SRResults}
\end{figure*}

\begin{figure*}[t]
    \centering
    \hspace*{\fill}
	\subfigure[$\text{CNN-BP}_{80}$.]{
	    \label{fig:LR_DeepCAM}
		\includegraphics[width=0.23\textwidth]{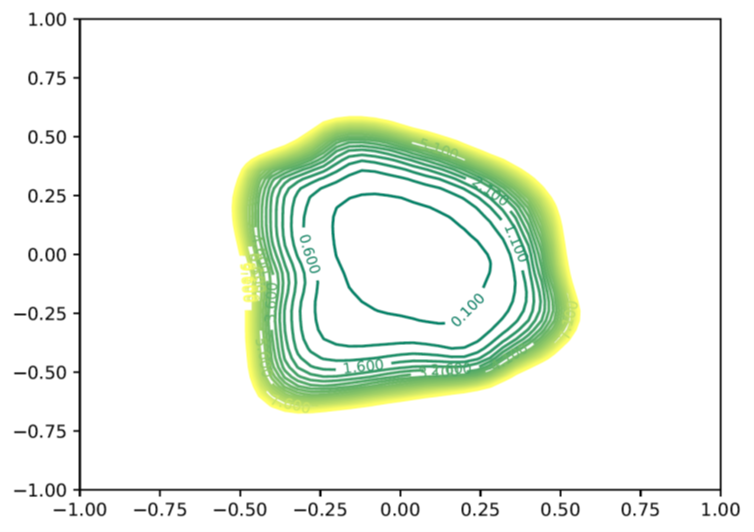}}
	\hfill
	\subfigure[$\text{CNN-BP}_{40}$.]{
	    \label{fig:DeepAM_DeepCAM}
		\includegraphics[width=0.23\textwidth]{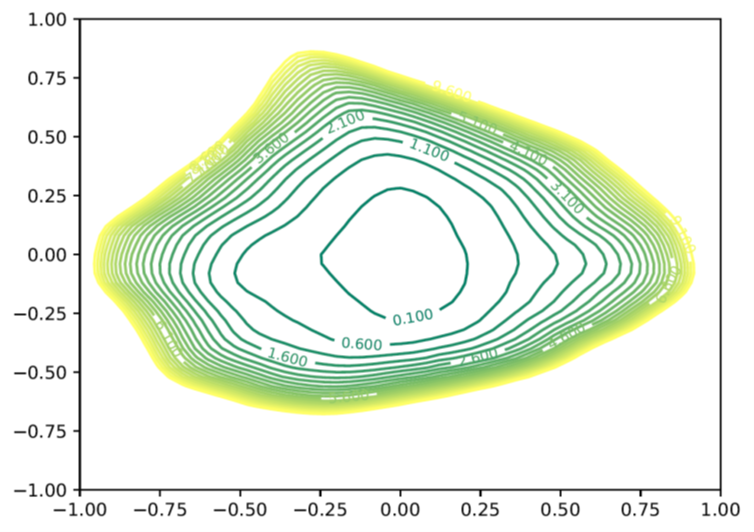}}
	\hfill
	\subfigure[DeepCAM.]{
	    \label{fig:DeepCAM_DeepCAM}
		\includegraphics[width=0.23\textwidth]{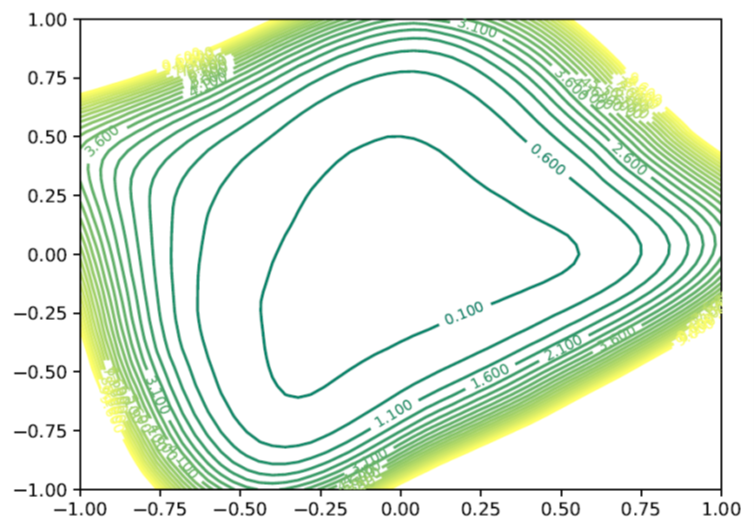}}
	\hfill
	\subfigure[$\text{DeepCAM}_{\text{bp}}$.]{
	    \label{fig:DeepCAM_DeepCAM}
		\includegraphics[width=0.23\textwidth]{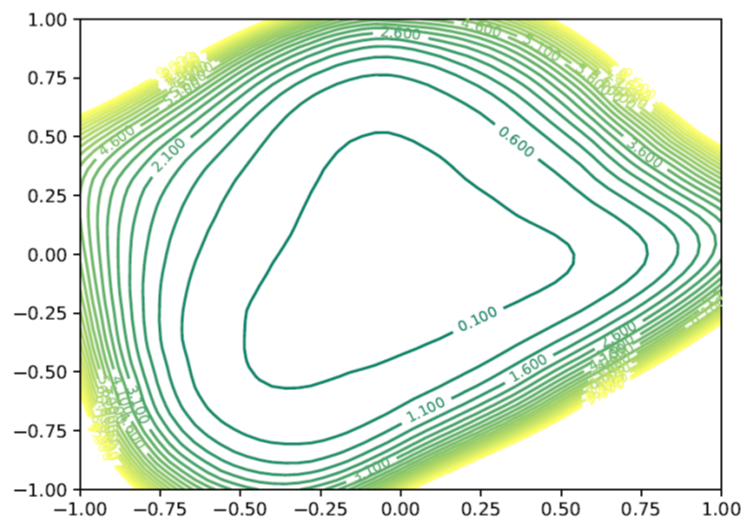}}
	\hspace*{\fill}
	
    \caption{Visualization the 2D surface of minima obtained with different methods. The sharpness of minimizers correlates well with generalization error. A wider, and flatter minimizer usually has better generalization ability. The minimizers of DeepCAM and $\text{DeepCAM}_{\text{bp}}$ are flatter and wider than that of $\text{CNN-BP}_{40}$ and $\text{CNN-BP}_{80}$.}
    \label{fig:DeepCAM_losssurface}
\end{figure*}

SC-based method \cite{zeyde2010single}, ANR \cite{timofte2013anchored} and A+ \cite{timofte2014a+} are patch-based. SC-based method \cite{zeyde2010single} is based on synthesis sparse representation and has a LR synthesis dictionary with 1024 atoms and a corresponding HR synthesis dictionary. The input feature is the compressed $1$-st and $2$-nd order derivatives of the image patch and is obtained using Principal Component Analysis (PCA). ANR and A+ \cite{timofte2013anchored, timofte2014a+} are based on clustering and assign each cluster a linear regression model. They use the same feature as in \cite{zeyde2010single} while require a huge number of free parameters. The SRCNN method \cite{dong2014learning} is based on a convolutional neural network with 2 convolutional layers of 64 and 32 filters. The spatial filter size is $9 \times 9$, $1 \times 1$ and $5 \times 5$, respectively.



DeepCAM used for comparison is a 3-layer DeepCAM. For the convolutional analysis dictionary, the spatial filter size is $3 \times 3$ at all layers and the filter number is 9, 25, 100 for layer 1, 2 and 3, respectively. 
The convolutional synthesis dictionary is with spatial filter size $5 \times 5$. Therefore, the effective filter size of DeepCAM is $11 \times 11$.

Table \ref{tab:NumParam} shows the number of free parameters in different single image super-resolution methods. The SC-based method \cite{zeyde2010single} requires a relatively small number of parameters which mainly comes from two synthesis dictionaries. The ANR method \cite{timofte2013anchored} and the A+ method \cite{timofte2014a+} have around 1 million free parameters because there are 1024 regressors with size $36 \times 28$. The SRCNN method \cite{dong2014learning} has the least number of parameters.
DeepAM has around 160,000 parameters. This is because the dictionaries are not structured and there are 3 layers of analysis dictionaries. The proposed DeepCAM has only approximately 35,000 free parameters since each structured convolutional dictionary shares a small number of free parameters though it has a huge size.

We denote with $\text{CNN-BP}_{DS}$ the deep neural network with the same structure as DeepCAM and trained using backpropagation algorithm \cite{rumelhart1985learning} with learning rate decay step $DS$ and total $5\times DS$ epochs for training. The implementation of DNNs is based on Pytorch with Adam optimizer \cite{kingma2014adam}, batch size $1$, initial learning rate $1\times10^{-3}$, and decay rate $0.1$. The training data has been arranged into $36 \times 36$ and $72 \times 72$ patch pairs.

Table \ref{tab:Set5_DeepCAM} shows the evaluation results of different methods on \textit{Set5}. DeepCAM outperforms SC-based method and ANR by around 0.5 dB, and has similar performance as SRCNN and DeepAM, while has around 0.2 dB lower PSNR than A+. The parameters setting of $\text{CNN-BP}_{40}$ has been tuned to achieve the best performance on \textit{Set5}.
$\text{CNN-BP}_{80}$ and $\text{CNN-BP}_{40}$ have been used for comparison and achieves around 0.1 dB and 0.2 dB higher PSNR than DeepCAM. $\text{DeepCAM}_{\text{bp}}$ represents the backpropagation fine-tuned version of DeepCAM with Adam optimizer \cite{kingma2014adam}, batch size $1$, initial learning rate $1\times10^{-4}$, and with total 20 epochs.
$\text{DeepCAM}_{\text{bp}}$ achieves better performance than DeepCAM. Its average PSNR is comparable to that of A+ and CNN-BP.

Table \ref{tab:Set14_DeepCAM} shows the evaluation results of different single image super-resolution methods on \textit{Set14} and results similar to those in Table \ref{tab:Set5_DeepCAM} can be observed.
The average PSNR of DeepCAM is around 0.3 dB higher than that of SC-based method and ANR, while it is around 0.15 dB lower than that of A+. DeepCAM achieves performance similar to SRCNN, DeepAM and $\text{CNN-BP}_{40}$. $\text{DeepCAM}_{\text{bp}}$ achieves improved performance than DeepCAM.

It is interesting to note that DeepCAM significantly outperforms $\text{CNN-BP}_{40}$ and $\text{CNN-BP}_{80}$ on “ppt3” and “zebra” which contain sharp edges with small scales. 
In particular, on “ppt3”, DeepCAM outperforms $\text{CNN-BP}_{40}$ and $\text{CNN-BP}_{80}$ by 0.4 dB and 1.6 dB, respectively.
Figure \ref{fig:DeepCAM_SRResults} shows the reconstructed HR images of the testing image “ppt3” using $\text{CNN-BP}_{80}$, $\text{CNN-BP}_{40}$ and DeepCAM. We can find that $\text{CNN-BP}_{80}$ and $\text{CNN-BP}_{40}$ cannot reconstruct well the characters. A possible reason is that $\text{CNN-BP}$ has a weaker generalization ability on the unseen testing data.

Li \textit{et al.} \cite{li2018visualizing} proposed a method to visualize the loss surface landscape around the minimizer of a deep model. The sharpness of the loss surface landscape of a minimizer is well correlated to the generalization ability. That is, a wider and flatter minimizer has better generalization ability. From Figure \ref{fig:DeepCAM_losssurface}, we can see that $\text{CNN-BP}_{80}$ has the sharpest and the most narrow 2D loss surface, $\text{CNN-BP}_{40}$ has a wider and flatter one but it is still not as wide as that of DeepCAM. We can also find that performing backpropagation fine-tuning on DeepCAM does not change significantly the 2D loss surface. The visualization in Figure \ref{fig:DeepCAM_losssurface} correlates well with the simulation results in Table \ref{tab:Set14_DeepCAM}. The reason that DeepCAM possesses a stronger generalization ability is probably due to our information preserving and clustering design.


\section{Conclusions}

In this paper, we proposed a convolutional analysis dictionary learning algorithm by exploiting the properties of the Toeplitz structure within the convolution matrices. The proposed algorithm can impose the global rank property on learned convolutional analysis dictionaries while performing learning on the low-dimensional signals. We then proposed a Deep Convolutional Analysis Dictionary Model (DeepCAM) framework which consists of multiple layers of convolutional analysis dictionary and soft-threshold pairs and a single layer of convolutional synthesis dictionary. Similar to DeepAM, the convolutional analysis dictionaries are designed to be made of an information preserving analysis dictionary (IPAD) and a clustering analysis dictionary (CAD). The IPAD preserves the information from the input image, while the CAD generates discriminative feature maps for image super-resolution.
Simulation results show that our proposed DeepCAM achieves comparable performance with other existing single image super-resolution methods while also having a good generalization capability.
\ifCLASSOPTIONcaptionsoff
  \newpage
\fi


\bibliographystyle{IEEEtran}
\bibliography{bibs}

\end{document}